\documentclass[10pt,twocolumn,letterpaper]{article}

\usepackage[pagenumbers]{cvpr} % To force page numbers, e.g. for an arXiv version

\def\name{\textsc{FNAC}\xspace}
\usepackage{url}
\usepackage{graphicx}
\usepackage{amsmath}
\usepackage{amssymb}
\usepackage{booktabs}
\usepackage{multirow}
\usepackage{tabularx}
\usepackage{array}
\usepackage{color}
\usepackage[pagebackref,breaklinks,colorlinks]{hyperref}

\usepackage[normalem]{ulem}

\usepackage[capitalize]{cleveref}
\crefname{section}{Sec.}{Secs.}
\Crefname{section}{Section}{Sections}
\Crefname{table}{Table}{Tables}
\crefname{table}{Tab.}{Tabs.}

\crefname{section}{Sec.}{Secs.}
\Crefname{section}{Section}{Sections}
\crefname{table}{Tab.}{Tabs.}
\Crefname{table}{Table}{Tables}
\crefname{figure}{Fig.}{Figs.}
\Crefname{figure}{Figure}{Figures}
\crefname{equation}{Eq.}{Eqs.}
\Crefname{equation}{Equation}{Equations}

\newcommand{\tofill}[1]{\textcolor{red}{[TO FILL]}}

\def\cvprPaperID{5862} % *** Enter the CVPR Paper ID here
\def\confName{CVPR}
\def\confYear{2023}

\def\methodone{\textsc{FNS}\xspace}
\def\methodtwo{\textsc{TNE}\xspace}

\begin{document}

%%%%%%%%% TITLE - PLEASE UPDATE
% \title{Learning Modal specific similarity for multi-modal self-supervised sound localization}
\title{Learning Audio-Visual Source Localization via \\ False Negative Aware Contrastive Learning}

\author{Weixuan Sun$^{1,5}$ \thanks{Indicates equal contribution} \ , Jiayi Zhang$^{2 \ *}$ , Jianyuan Wang$^{3}$, Zheyuan Liu$^{1}$, 
Yiran Zhong$^{4}$, \\Tianpeng Feng$^{5}$, Yandong Guo$^{5}$, Yanhao Zhang$^{5}$, Nick Barnes$^{1}$ \\
\\ 
% \vspace{-3mm}
$^{1}$Australian National University, $^{2}$Beihang University, $^{3}$The University of Oxford,  \\
$^{4}$Shanghai AI Lab, $^{5}$OPPO Research Institute.
% {\tt\small firstauthor@i1.org}
% For a paper whose authors are all at the same institution,
% omit the following lines up until the closing ``}''.
% Additional authors and addresses can be added with ``\and'',
% just like the second author.
% To save space, use either the email address or home page, not both
% \and
% Second Author\\
% Institution2\\
% First line of institution2 address\\
% {\tt\small secondauthor@i2.org}
% }
\vspace{-5mm}
}
\maketitle

%%%%%%%%% ABSTRACT
\begin{abstract}
Self-supervised audio-visual source localization aims to locate sound-source objects in video frames without extra annotations. 
Recent methods often approach this goal with the help of contrastive learning, which assumes only the audio and visual contents from the same video are positive samples for each other. 
However, this assumption would suffer from false negative samples in real-world training.
For example, for an audio sample, treating the frames from the same audio class as negative samples may mislead the model and therefore harm the learned representations (\textit{e.g.}, the audio of a siren wailing may reasonably correspond to the ambulances in multiple images).  
Based on this observation, we propose a new learning strategy named False Negative Aware Contrastive (\name) to mitigate the problem of misleading the training with such false negative samples. 
Specifically, we utilize the intra-modal similarities to identify potentially similar samples and construct corresponding adjacency matrices to guide contrastive learning. 
Further, we propose to strengthen the role of true negative samples by explicitly leveraging the visual features of sound sources to facilitate the differentiation of authentic sounding source regions. 
\name achieves state-of-the-art performances on Flickr-SoundNet, VGG-Sound, and AVSBench, which demonstrates the effectiveness of our method in mitigating the false negative issue. 
The code is available at \url{https://github.com/OpenNLPLab/FNAC_AVL}. 

\end{abstract}

%%%%%%%%% BODY TEXT
\section{Introduction}
\label{section intro}
When hearing a sound, humans can naturally imagine the visual appearance of the source objects and locate them in the scene. 
This demonstrates that audio-visual correspondence is an important ability for scene understanding. 
Given that unlimited paired audio-visual data exists in nature, there is an emerging interest in developing multi-modal systems with audio-visual understanding ability.
Various audio-visual tasks have been studied, including sound source localization~\cite{mo2022localizing,chen2021localizing,qian2020multiple,senocak2018learning,liu2022exploiting,lin2021unsupervised, ramaswamy2020see}, audio-visual event localization~\cite{tian2018audio,zhou2021positive,yu2022mm,wu2019dual}, audio-visual video parsing~\cite{lin2021exploring,cheng2022joint,tian2020unified} and audio-visual segmentation~\cite{zhou2022avs,zhou2023avss}.
In this work, we focus on unsupervised visual sound source localization, with the aim of localizing the sound-source objects in an image using its paired audio clip, but without relying on any manual annotations. 
% \NB{sounding objects is a bit strange. Maybe sound-source objects - makes sense and corresponds with the idea of sound-source localization, and extending this to object localization}

 \begin{figure}[!t]
   \begin{center}
   {\includegraphics[width=0.9\linewidth]{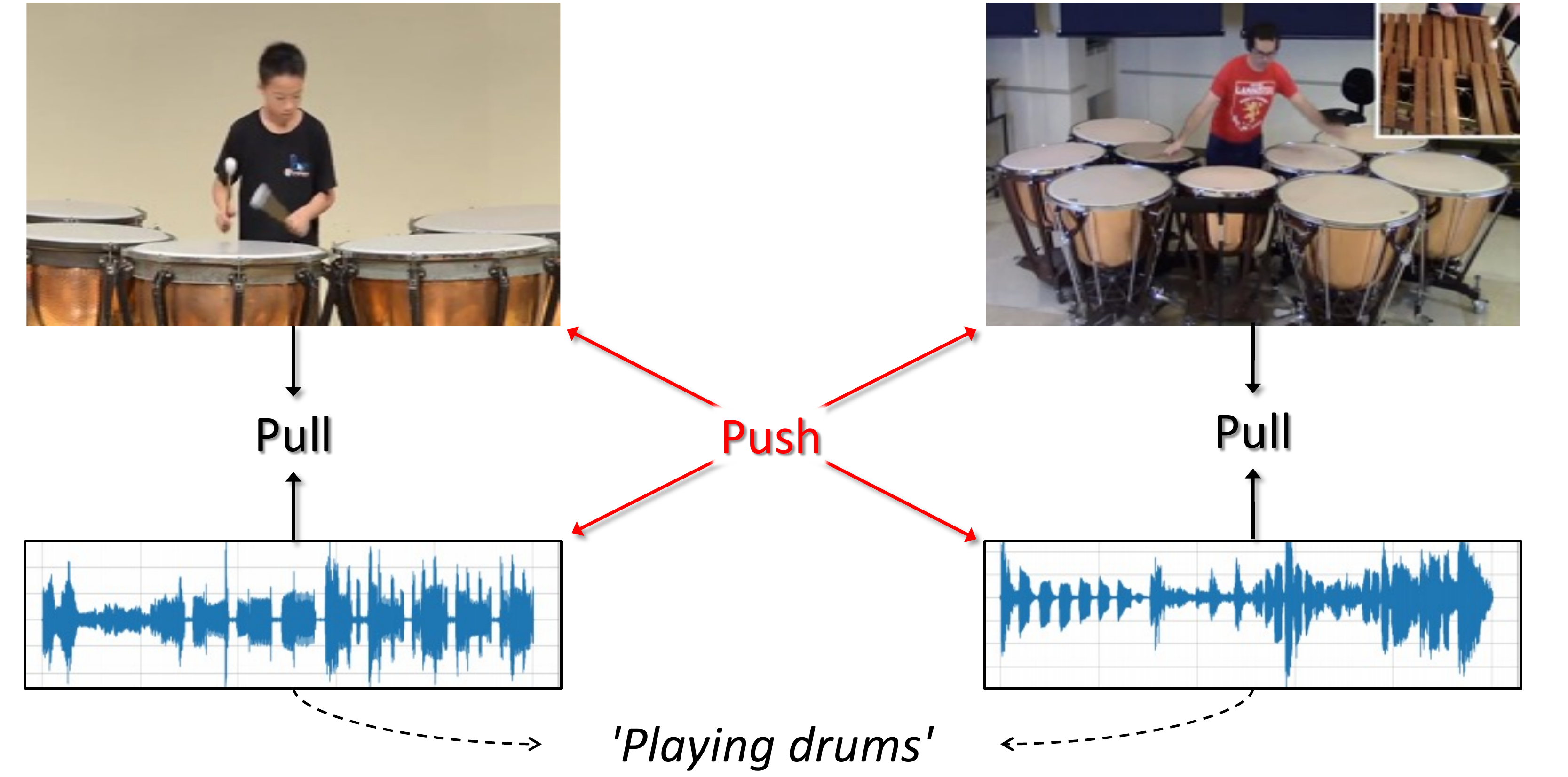}} 
   \end{center}
   \vspace{-1mm}
\caption{\textbf{False negative in audio-visual contrastive learning.} Audio-visual pairs with similar contents are falsely considered as negative samples to each other and pushed apart in the shared latent space, which we find would affect the model performance. 
% \NB{Should you say something like: FNAC corrects this by ..}
}
\vspace{-2mm}
   \label{fig:intro}
\end{figure}

% Unsupervised visual sound source localization is a challenging task. 
%
% To understand the audio-visual pairs without any human annotations, it is essential to leverage the corresponding relationships between the audio and the paired image.
% 
The essence of unsupervised visual sound source localization is to leverage the co-occurrences between an audio clip and its corresponding image to extract representations. 
A major part of existing methods~\cite{mo2022localizing,chen2021localizing,liu2022exploiting,senocak2018learning,lin2021unsupervised} formulates this task as contrastive learning.
% An audio clip and its paired image are considered as two different views of a joint embedding.
For each image sample, its paired audio clip is viewed as the positive sample, while \textit{all other} audio clips are considered as negative.
Likewise, each audio clip considers its paired image as positive and \textit{all others} as negative.
% \NB{not clear what vice versa is referring to here? is it necessary?}.
As such, the Noise Contrastive Estimation (NCE) loss~\cite{oord2018representation,tian2020contrastive} is used to perform instance discrimination by pushing closer the distance between a positive audio-image pair, while pulling away any negative pairs.
% Thus, paired audio-visual features are encouraged to be close to each other.
% \sout{During the inference stage, the localization results are obtained by calculating the similarity between the audio and image representations.}
%
% \textcolor{blue}{(However, such scheme asumes ..., which inevitably leads to ..., i.e, specify the concrete category corresponding to Fig.1(a))}
However, the contrastive learning scheme above suffers from the issue of false negatives during training, \textit{i.e.}, audio/image samples that belong to the semantically-matched class but are not regarded as a positive pair (due to the lack of manual labeling).
% appearing in the same batch may have close visual/audio content but are regarded as negatives.
% \sout{and pushed apart}
% \jy{not sure if class conflict issue is the best name; possibly just call it false negative problem?} 
A typical example is shown in Fig.~\ref{fig:intro}.
% For example, semantically matched audio-visual pairs are considered as negative to each other and falsely pushed apart, as seen in Fig.~\ref{fig:intro}.
% As depicted in Fig.~\ref{fig:intro}, the two sounds of \textit{playing drums} are semantically-matched and  should share equal roles in locating the drums in visual scenes, but they are falsely pushed apart.
% As shown in Fig.~\ref{fig:intro}, the two audio-visual samples share the
% the first and third samples will be treated as negatives by the anchor in a typical contrastive framework. However, the first sample and the anchor share similar audio signals (same audio class, \textit{i.e.}, playing drums). 
% \jiayi{They share equal roles in locating the soundingobject, drums.}
% Thus, the first sample is a false negative.
Research shows~\cite{arora2019theoretical,zheng2021weakly,senocak2022learning,huynh2022boosting} that these false negatives will lead to contradictory objectives and harm the representation learning.
% a typical contrastive framework will treat the first and second as both negative samples of $A_1$, but the sounds of the first and the second samples may be quite close. 
% have the same semantic class which is 'playing drum'.
% \jiayi{Perhaps the example in Fig.1 can be described more clearly} 

%
% harm the representation learning as the model fails to discriminate hard positive samples.
%
% We first analyze this problem in .

% To validate 
% To validate our hypothesis, we first analyze the statistical numbers of false negative samples in real-world training. 
% To verify our hypothesis, w
Motivated by this observation, we assess the impact of false negatives in real-world training. 
We discover that with a batch size of 128, around $40\%$ of the samples in VGG-Sound~\cite{chen2020vggsound} will encounter at least one false negative sample during training.
% As expected, 
% On the widely used benchmark VGG-Sound 144k~\cite{chen2020vggsound}, we found that around $40\%$ samples have encountered at least one false negative sample (batch size 128), 
% and the average false negative ratio is .... \tofill{}.
% \wx{isn't this same as 40 \%?} \jy{40\% samples have negative samples, while one sample could have n negative samples.}
%
We then validate that false negatives indeed harm performance by artificially increasing the proportion of false negatives during training, and observing a noticeable performance drop.
% In addition, if we artificially increase the proportion of false negatives during training, a noticeable performance drop is observed, which validates the harm of the false negatives.
% on the representation learning.  
% \jiayi{Hi Weixuan, could we move the following paragraph into the related work to simplify the introduction?}
% % As far as we know, there is just one work that identifies similar problem. 
% The only work that studies a similar problem is ~\cite{senocak2022learning}, which tried to find potential class-matched samples with the localization maps, since audio clips that belong to the same semantic class will localize the same areas in a picture. 
% % since same-class samples generate similar localization results to the original pairs. \jy{I am a bit confused about the description of this related work. Possibly we could polish it a bit and make it clearer} 
% However, it still requires comparison of inter-modal localization maps between all possible pairs.
% \jiayi{I'm confused about the statement. Discuss later?
To make matters worse, larger batch sizes are often preferred in contrastive learning~\cite{oord2018representation}, but it may inadvertently increase the number of false negative sample pairs during training and affect audio-visual representation quality.
To this end, we propose a false-negative aware audio-visual contrastive learning framework (\name), where we employ the intra-modal similarities as weak supervision. 
% We empirically find that the intra-modal similar samples are more likely from the semantically-matched class and should be considered as false negative samples in the inter-modal training.
Specifically, we compute pair-wise similarities between all audio clips in a mini-batch without considering the visual to form an audio intra-modal adjacency matrix.
% we calculate a intra-modal 
% adjacency matrix which measures the pair-wise similarities across all audio clips in the current mini batch. 
Likewise, in the visual modality, we obtain an image adjacency matrix. 
% \ZL{you often use the word symmetrically, but its meaning here is a bit off.}
We found that the adjacency matrices effectively identify potential samples of the same class within each modality (Fig.~\ref{fig:similar samples}). The information can then be used to mitigate the false negatives and enhance the effect of true pairings.

Specifically, we propose two complementary strategies: 1) \textbf{\methodone} for False Negatives Suppression, and 2) \textbf{\methodtwo} for True Negatives Enhancement.
% Feature similarities are often investigated to mine similar samples in single modality~\cite{zheng2021weakly,chuang2020debiased,caron2018deep}. 
% and which is more robust than cross modalities.
% are more robust in single modality than cross modalities
% Thus, we assume that the
% \jiayi{The assumption here seems a little strong. For example, false negatives are those samples highly semantically similar, while samples from the same class are just more likely to be similar. Maybe we could put it another way?}
% During the contrastive learning, the main supervision signal is the audio-visual pairing in which we compute a inter-modal adjcency matrix. 
% Specifically, we introduce two techniques to utilize the intra-modal similarities.
% In our strategy,
% Our strategy deals with the false negative problem by introducing two . 
% \wx{todo make this clear}
First, when optimizing the NCE loss, \methodone regularizes the inter-modal and intra-modal similarities. 
% \jiayi{The following sentence is a little confusing.}
Intrinsically, 
intra-modal adjacency explores potential false negatives by the similarity intensities and the pulling forces applied to these false negatives are canceled accordingly.
% Intrinsically, similar samples in each modality are less considered in NCE, so the pulling forces applied on these potential false negatives are canceled according to the intra-modal similarity intensity. 
% \ZL{What does it mean?}
% Consequently, the misleading effects are suppressed.
% which suppresses the contribution of the potential false negatives according to the similarity intensity of the adjacency matrices. 
%
% therefore the misleading contribution from the potential false negatives is moderately suppressed according to the similarity intensity of the adjacency matrices. 
% In practice, 
% we keep the NCE loss unchanged and directly regularize the distances between the inter-modal adjacency matrix and intra-modal adjacency matrices, which prevents false negatives to be pushed away from the anchor. 
% Thus, only the true negative samples are pushed apart.
Furthermore, we introduce \methodtwo to emphasize the true negative influences in a region-wise manner, which in turn reduces the effect of false negative samples as well.
We adopt the audio adjacency matrix to identify dissimilar samples, i.e., true negatives. 
Intuitively, dissimilar (true negative) sounds correspond to distinct regions, 
so the localized regions across the identified true negatives are regularized to be different.
% Such mechanism encourages the model to localize discriminative regions 
% we extend the intra-modal similarities to guide region-level contrastive learning. 
% Intuitively, similar audio clips should correspond to similar visual appearances, and vice versa.
% We first obtain the localized image regions from every audio-visual pairs.
% Then regularize the adjacency across the localized regions to align with the audio adjacency matrix. 
Such a mechanism encourages the model to discriminate genuine sound-source regions and suppress the co-occurring quiet objects.
% In this way, we 
% Besides, we can further expand the intra-modal similarity guidance to region-level learning, such that the model is encouraged to identify true sounding locations.
% Intuitively, similar audios should have similar localized visual appearance, and vice versa.
% In this way, we obtain each pair's localization results and 
% encourage the similarities between the localized visual features align with the audio similarities. Thus, the model would discriminate the true sounding regions from the co-occurring objects or background. 
% Specifically, we compute intra-modal adjacency matrices of audio and visual respectively, then use the matrices to guide the contrastive learning.
% The contrastive learning framework provides robust intra-model similarities, i.e., for each image we can find its nearby samples by only considering the image representations and it also applies to audio.
% We find that each modal provides robust intrinsic intra-modal similarities, which can be leveraged to guide the cross-modal contrastive learning.
% Our proposed framework simultaneously address two types of hard positive issues in audio-visual representation learning.
% Specifically, we calculate a batch-wise adjacency matrix in audio modal, which encodes pair-wise similarities between all audios.
we conduct extensive analysis to demonstrate the effectiveness of our proposed method and report competitive performances across different settings and datasets.
% We conduct extensive experiments to demonstrate the effectiveness of our proposed method across different settings and several datasets.

In summary, our main contributions are:
\begin{itemize}
    \item We investigate the false negative issue in audio-visual contrastive learning. We quantitatively validate that this issue occurs and harms the representation quality. 
    \item We exploit intra-modal similarities to identify potential false negatives and introduce \methodone to suppress their impact.
    \item We propose \methodtwo, which emphasizes true negatives using different localization results between the identified true negatives, thus encouraging more discriminative sound source localizations.
    % , which .
    % \NB{the wording of this last point seems not quite clear}
    % \NB{adopts seems a weak word here - can you be more desciptive of the relationship between the intra-modal and the other.} 
    % the intra-modal adjacency to enhance the true negative contribution by region-wise contrastive \NB{loss?}. \NB{Could this be a bit sharper pointing to novelty and contribution a bit more specifically}
\end{itemize}

% \jy{Do we need to highlight that, we assume the samples with the same audio class tend to be False Negative. I noticed some sentences like "they share the same audio class, i.e., are false negative to each other". Possibly, such a statement is not quite accurate. }
% \jiayi{I share a common concern about it. At the same time, I think it should be emphasized that this assumption is just for the sound source location task.}

% \NB{Is the intro a little bit long?}

\section{Related Work}
\label{section related}

\paragraph{False Negatives in Contrastive Learning.}
Typical contrastive learning employs instance discrimination~\cite{wu2018unsupervised} as a pretext task, in which two augmented views of the same image are considered as positive pairs, while views of all other images are treated as negative pairs, regardless of semantic similarities. 
Such a scheme inevitably suffers from the False Negatives issue~\cite{zheng2021weakly,dwibedi2021little,huynh2022boosting,chen2021incremental}, which indicates that instances sharing the same semantic concepts are falsely treated as negatives, thus misleading model learning. 
%Motivated by this observation, some methods are proposed to tackle this issue.
Based on this, some works attempt to incorporate similar instances into model training to eliminate the impact of false negatives.
For example, Zheng \textit{et al.}~\cite{zheng2021weakly} model a nearest neighbor graph for each batch of instances and execute a KNN-based multi-crop strategy to detect false negatives. 
% Zheng \textit{et al.}~\cite{zheng2021weakly} assume similar instances share the same weak label and are expected to be aggregated in the latent space. They model a nearest neighbor graph for each batch of instances and perform a KNN-based multi-crop strategy to detect more positives for each weak label.
Similarly, Dwibedi \textit{et al.}~\cite{dwibedi2021little} sample nearest neighbors from the dataset and treat them as positives for contrastive learning. 
% Other works adopt clustering-based methods to encode semantic structures~\cite{asano1911self,caron2018deep,li2020prototypical}. 
% For example, Li \textit{et. al}~\cite{li2020prototypical} assign several prototypes for different groups of semantically similar instances and construct a contrastive loss which encourages an instance to get closer to its corresponding prototype in latent space.
% Some methods are proposed on the false negative issues in the single-modal contrastive learning.
More recently, \cite{huynh2022boosting} and \cite{chen2021incremental} study how to identify false negatives without class labels and explicitly remove detected false negatives by two strategies, elimination and attraction, to improve contrastive loss.
Other works use clustering-based methods to encode semantic structures~\cite{asano1911self,caron2018deep,li2020prototypical} and then perform contrastive learning on these semantically similar cluster centers. 
In this paper, we propose and explore a similar problem in self-supervised audio-visual learning.

% [A] Pedro Morgado, Ishan Misra, and Nuno Vasconcelos, Robust Audio-Visual Instance Discrimination

% [B] Pedro Morgado, Nuno Vasconcelos and Ishan Misra, Audio-Visual Instance Discrimination with Cross-Modal Agreement

% [C] Arda Senocak, Hyeonggon Ryu, Junsik Kim and In So Kweon, Learning sound localization better from semantically similar samples

\paragraph{Self-Supervised Sound Source Localization.} 
% When we hear a dog barking, we could see the dog simultaneously.
Sound source localization aims to learn to locate sound-source regions in videos.
% by exploiting the natural co-occurrence of visual and auditory cues. 
Recent approaches extensively leverage contrastive learning based on audiovisual correspondence to address this issue. 
For example, ~\cite{arandjelovic2017look,arandjelovic2018objects,senocak2018learning} adopt a dual-stream architecture to extract unimodal features respectively and then calculate a contrastive loss to update the audiovisual network. The final localization map is usually obtained by calculating the cosine similarity between audio and visual features. 
Following the paradigm, 
Mo \textit{et al.}~\cite{mo2022localizing} further propose an object-guided localization (OGL) module, an extra pre-trained visual encoder, to introduce visual priors into the localization results. 
% The correspondence assumes that audio and visual signals from the same video are regularly aligned. However, unconstrained videos in real world may contain a lot of visual contents irrelevant the sound source.
% However, the aforementioned contrastive framework based on instance discrimination assumes that the paired audio-visual signals are regularly aligned and all mismatched samples are heterogeneous, specifically, it regards the paired image as positive and all other images in the same batch as negatives and vice versa.
Nevertheless, these works assume that the paired audio-visual signals are regularly aligned and all mismatched samples are heterogeneous. As aforementioned, the assumption ignores the semantic similarities between samples.
%! Consider whether to illustrate the significance to SSL with examples
% To reduce the impact of the correspondence assumption 
% , Hu \textit{et al.}~\cite{hu2019deep} first utilize the cluster centers of audio and visual representations for contrastive learning.
Accordingly, Chen \textit{et al.}~\cite{chen2021localizing} incorporate explicitly background regions with low correlation to the given audio into the framework and regard them as \textit{hard negatives}.
% Senocak \textit{et al.}~\cite{senocak2022learning} suppose the unmatched pairs may contain semantically matched audio-visual information and integrate these \textit{hard postives} into the training procedure, which is a similar problem to us.
In a slightly different task setting of audio-visual instance discrimination, false negative issues are also investigated.
~\cite{morgado2021audio} detects false negatives by defining sets of positive and negative samples via cross-modal agreement, where positive samples are defined as similar samples in both modalities.
~\cite{morgado2021robust} considers both false positive and false negative issues. 
% They model the similarity scores as a normal distribution and the outliers are down-weighted in the loss to reduce false positive contributions. 
For false negatives, they estimate the similarity across instances to compute a soft target distribution over instances so the contributions of false negatives are down-weighted.
One work that explores a similar problem to us is \cite{senocak2022learning}, which treats top-K semantically similar audio-visual pairs as \textit{hard positives} and explicitly integrates them into the contrastive loss.
Unfortunately, this method relies on manually selected hypermeters K and a hard threshold. 
% \wx{One work that explores a similar problem to us is ~\cite{senocak2018learning}, which adopts similar cross-modal localization maps to detect class-matched samples and regard them as false negatives in the training procedure. However, they require to compare inter-modal localization maps between all possible pairs. Differently, we .....}
% \jiayi{Hi Weixuan, both \cite{morgado2021audio} and \cite{morgado2021robust} explore a similar problem to ours, that is, False Negatives (\& Positives) in audio-visual contrastive learning. Although they investigate a general task AVID, I think we should discuss how our approach differs from theirs. For example, \cite{morgado2021audio} calculate the cross-modal agreement score between the anchor and other samples, which measures the semantic similarity between two samples. Then they select the top-K samples as the positive set of the anchor for contrastive learning. This method is similar to \cite{morgado2021audio}, both of which are hyperparameter sensitive.}
Unlike the previous methods, this paper uses intra-modal adjacency matrices to adaptively detect \textit{false negatives} and eliminate their impact.

\section{Method}
% In this section, we first revisit contrastive learning in the audio-visual localization task and present proof-of-concept experiments to investigate the false negative issue. 
% Then we introduce our xxxx framework (\name) to mitigate the problem and improve the performance.

%
% The overview of \name is illustrated in Fig \tofill{}.
%
% We adopt the intra-modal similarities to guide inter-modal learning.

% Background --
% Hypothesis --
% Hypothesis Verification --
% %
% Method:
% %
% There are two possible directions to mitigate the problem:
% 1. Identify false negative samples and reduce the role of false negative.
% 2. Raise the role of True negative . discriminate regions between true negative
% %
% sample-based -> mitigating false negative
% %
% region-based -> emphasizing true negative

In this section, we first investigate the problem raised by false negative samples in ~\cref{subsec:problem_def}, and then propose to mitigate the problem.
From two aspects, 
without supervision we
identify the false negative samples to explicitly suppress them in training~(\cref{subsec:suppress}), and use region-wise comparing learning to enhance the roles of true negative samples which hence relatively suppresses false negatives~(\cref{subsec:enhance}).
% we suppress the false positives by explicitly identifying them (\cref{subsec:suppress}) or implicitly .

% we suppress the false positives directly  
% Specifically, 

\subsection{Revisiting Audio-visual Contrasting Learning}\label{subsec:problem_def}
% \label{method_def}

\begin{figure}[t]
  \centering
  %\fbox{\rule{0pt}{2in} \rule{0.9\linewidth}{0pt}}
   \includegraphics[width=1\linewidth]{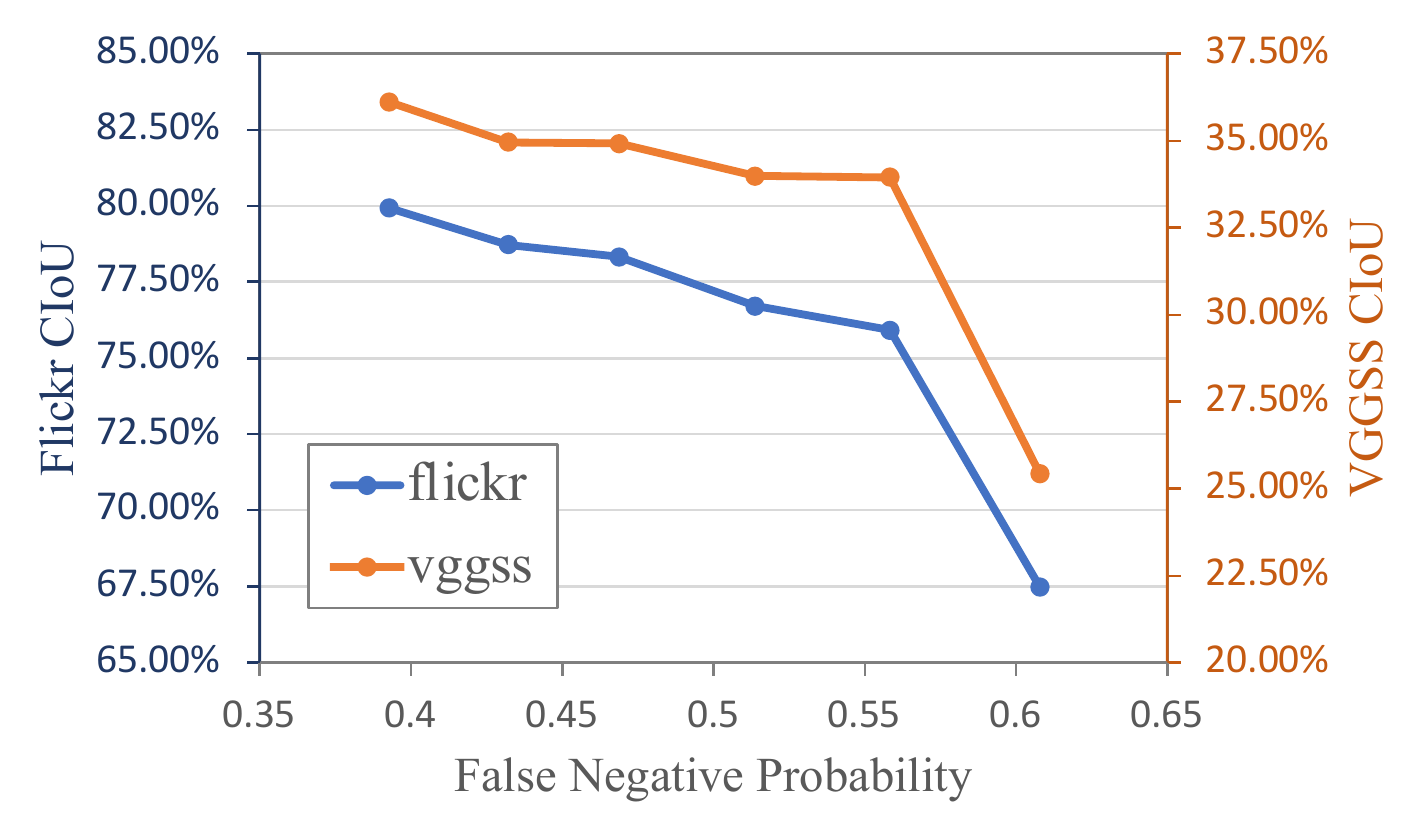}
   \caption{\textbf{Impact of false negatives on audio-visual representation learning}. 
   We adopt Consensus Intersection over Union (CIoU)~\cite{senocak2018learning} as an evaluation metric (higher is better) and report results on FlickrSoundNet~\cite{aytar2016soundnet} and VGG-SS~\cite{chen2020vggsound} test sets, depicted by \textcolor[RGB]{61,89,171}{blue} and \textcolor[RGB]{210,105,30}{brown}, respectively. 
   % \textcolor{navy}{blue} and \textcolor{orangered}{orange}, respectively. 
   % The model is trained on VGGSound-10k and evaluated on both FlickrSondNet~\cite{aytar2016soundnet} and VGG-SS~\cite{chen2020vggsound}. 
   An obvious performance decline is observed as the proportion of false negative samples increases. 
   % \jy{a quick question, vggss is lowercase in the figure while capital letter in left and right. Is it designed? And, do we need to mention the left and right axes are in different coordinates?} 
   }
   \label{fig:pilot}
\end{figure}

\begin{figure*}[!t]
   \begin{center}
   {\includegraphics[trim={0 3mm 0 0},clip,width=0.65\linewidth]{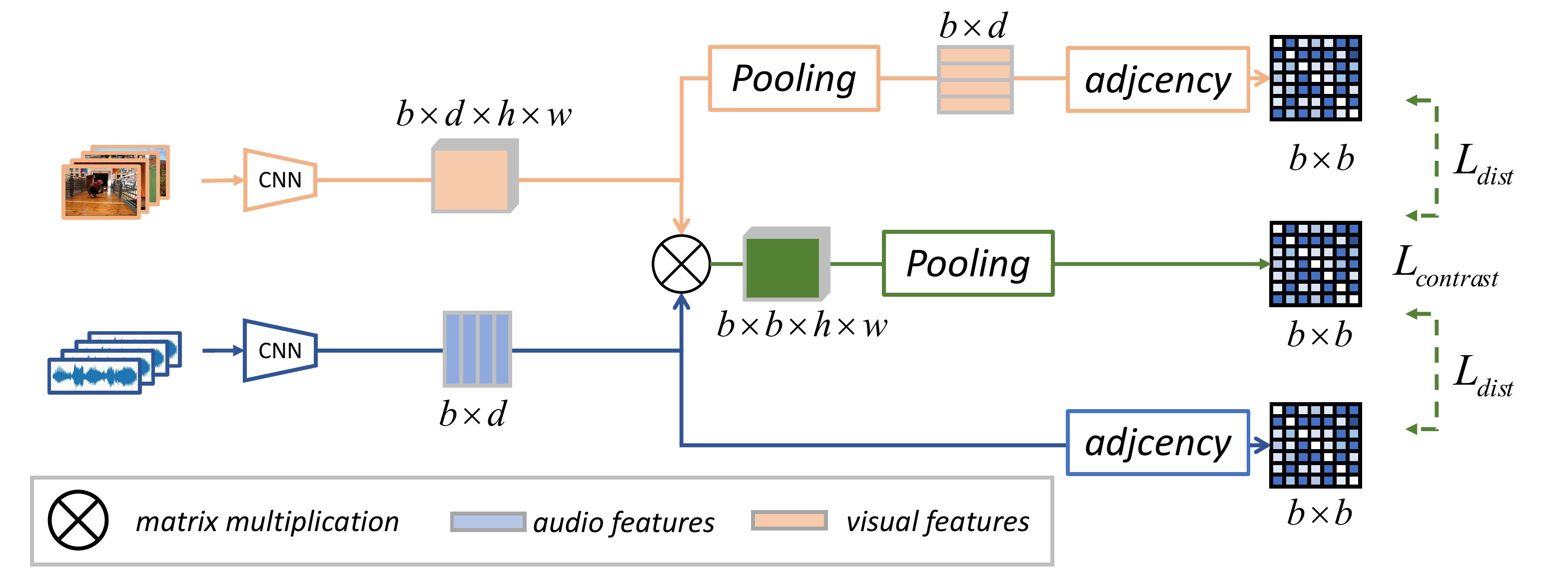}} 
   \end{center}
\caption{\textbf{An overview of False Negative Suppression (\methodone).} The main optimization objective is NCE~\cite{oord2018representation} loss on audio-visual pairs. The audio adjacency matrix and visual adjacency matrix are respectively constructed to suppress false negatives by regularizing NCE loss.}
% \vspace{-3mm}
   \label{fig:overview}
\end{figure*}

\begin{figure*}[!t]
   \begin{center}
   {\includegraphics[width=.7\linewidth]{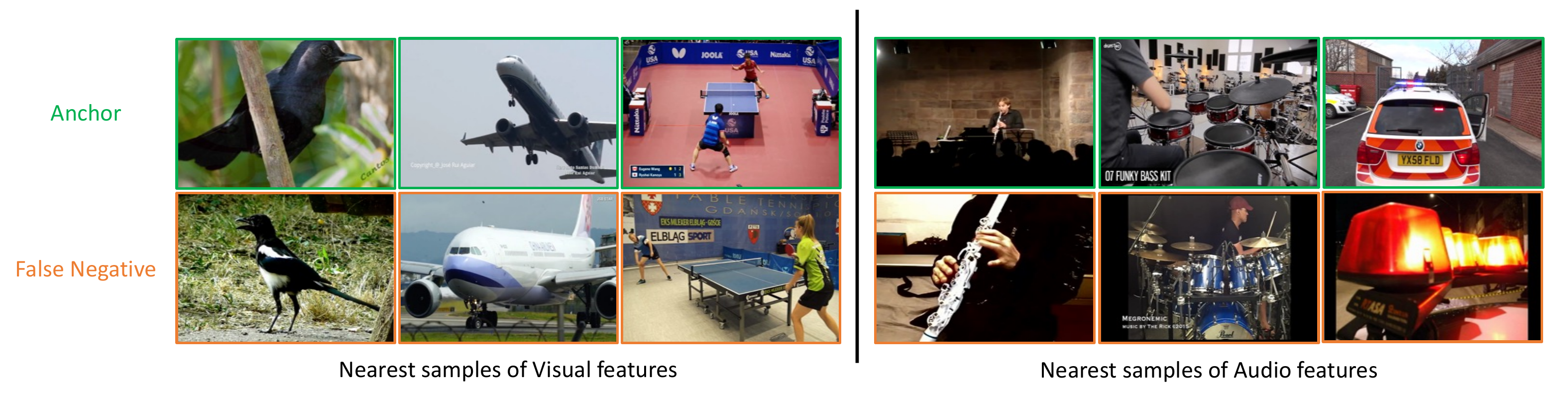}} 
   \end{center}
  \vspace{-4mm}
\caption{\textbf{Qualitative samples of the found potential false negatives in visual and audio modalities.} We report the \textcolor[RGB]{255,125,64}{nearest sample} for every \textcolor[RGB]{34,139,34}{anchor}. For the audio modality, we show the corresponding images here.}
   \label{fig:similar samples}
\end{figure*}

Various methods for audio-visual localization~\cite{mo2022localizing,senocak2018learning,lin2021exploring} employ a contrastive learning framework, in a manner of instance discrimination. 
Arguably, it assumes the audio-visual scenes from different videos are distinctive.
Therefore the paired audio and image from the same video are considered a positive pair while the samples (audio or image) from other pairs are regarded as negatives.
Specifically, we denote an audio-visual dataset as $\mathcal{D} \in \{(a_i, v_i), i \in [0, n)\}$, where $(a_i,v_i)$ represents a sample with paired audio-visual content. Usually, a two-stream network is used to encode audio and visual signals and then map them into a shared latent space. The audio and visual representations extracted from $(a_i, v_i)$ are denoted as $Z^a_i \in \mathbb{R}^{1 \times d}$ and $Z^v_i \in \mathbb{R}^{1 \times d}$ with $d$ as feature dimension.
Ideally, $Z^a_i$ and $Z^v_i$ represent the same semantic concept from the visual and audio perspectives, \textit{e.g.}, playing the drum.
The optimization objective of contrastive learning is to maximize the similarity between audio and visual representations from the same video while minimizing the similarity between features from different videos. 
Mathematically, it can be formulated in a modal-symmetric way for a pair $(a_i, v_i)$ as:
\begin{eqnarray}
\label{equation contra}
    \mathcal{L}_{\text{contrast\_i}} = - \log \frac{ \exp \left[ \frac{1}{\tau} \text{sim}(Z^a_i,Z^v_i) \right]  }{ \sum^{b}_{j} \exp \left[ \frac{1}{\tau} \text{sim}(Z^a_i, Z^v_j) \right] }  \nonumber \\
    - \log \frac{ \exp \left[  \frac{1}{\tau} \text{sim}(Z^v_i,Z^a_i) \right] }{ \sum^{b}_{j} \exp \left[  \frac{1}{\tau} \text{sim}(Z^v_i, Z^a_j) \right] },
\label{equation nce loss}
\end{eqnarray}
% Ideally, $v_i$ is supposed to be the only image 
% % \NB{Image or image/video? This was not clear in the intro} 
% that can produce the audio clip $a_i$, while all the other images (\textit{e.g.}, $v_j$) in the same mini-batch are considered as negative samples of $a_i$. 
% %
% In practice, the audio and visual signals are encoded with a two-stream network to obtain the batch-wise audio feature $Z^a \in \mathbb{R}^{b \times d}$ and visual feature  $Z^v \in \mathbb{R}^{b \times d}$ with $b$ as the batch size and $d$ as the feature dimension.  
% With similarity measured by cosine similarity or dot product, 
% a loss function called Noise Contrastive Estimation (NCE) loss is optimized in a modal-symmetric way:
% \begin{eqnarray}
% \label{equation contra}
%     L_{\text{contrast\_i}} = -\text{log} \frac{\frac{1}{\tau} sim(Z^a_i,Z^v_i)}{\sum^{b}_{j} \text{exp}(\frac{1}{\tau} sim(Z^a_i, Z^v_j) ) }  \nonumber \\
%     -\text{log} \frac{\frac{1}{\tau} sim(Z^v_i, Z^a_i)}{\sum^{b}_{j} \text{exp}(\frac{1}{\tau} sim(Z^v_i, Z^a_j) )} \label{equation nce loss}
% \end{eqnarray}

% \jiayi{Hi Weixuan, there is a problem with the formula (1) and (2). The condition j != i in the Sum formula of the denominator should be deleted, because the fraction here actually represents the SoftMax function.}
\noindent
where $\tau$ is a temperature hyper-parameter, $b$ denotes batch size and $\text{sim}$ represents the similarity function. 
Intuitively, this loss implies that each audio feature $Z^a_i$ is pushed close to its paired visual feature $Z^v_i$ in the shared latent space, while pulled apart from the rest $(b-1)$ visual features $Z^v_{j, j \neq i}$.
However, as discussed, there exist $\hat{Z^a_j}$ and/or $\hat{Z^v_j}$ that are semantically similar to $(Z^a_i, Z^v_i)$, \textit{i.e.}, false negatives. 
% For instance, the audios of (1) and (2) in Fig~\ref{fig:intro} are both playing drums and are similar. 
In this situation, forcing $\text{sim}(Z^a_i, \hat{Z^v_j})$ or $\text{sim}(Z^v_i, \hat{Z^a_j})$  to be small might perplex the model training and lead to a non-optimal representation.
% \jy{a double-check, false negatives can be both visual and audio right?} \jiayi{Yes, both are right.}

% Intuitively, this loss is a $b$ way softmax classifier.
% Each audio sample $a_i$ is pulled to its paired image $v_i$ to approach a similar representation in the latent space and pushed apart from other image samples.
% However, in real-world training, there exist false negative samples like $\hat{Z}_{j, j \neq i}$.
% For example, the audio signal $a_j$ may be quite close to $a_i$ (\textit{e.g.}, both siren wail), though the visual contents are totally different.
% In this situation, pushing $Z^v_i$ and $Z^a_j$ away may lead to a non-optimal representation.

% of $\hat{Z}_{j, j \neq i}$ may be quite close to that of $\hat{Z}_{i}$
% However, the false negative samples $\hat{Z}_{j, j \neq i}$ exist in the batch but are falsely pushed away from the class-matched sample.

\paragraph{Impact of False Negatives.}
Based on this observation, we conduct several pilot experiments to verify the issue of false negatives and their influence on audio-visual representation learning. 
For simplification, we reasonably assume that samples that share the same manually labeled category are False Negatives.
% \textit{To simplify the study, we assume the samples that share the same manually labeled category tend to be False Negatives.}
% We provide several pilot tests to validate the class conflict issue and its influence on audio-visual representation learning.
% Specifically, we run trials with a batch size of $128$ on the VGG-Sound 10K dataset~\cite{chen2020vggsound}, which has $309$ different categories.
%
Firstly, we examine the data distribution during the training procedure.
We find that over $\mathbf{39.27\%}$ samples suffer from at least one false negative sample when training with a batch size of 128 on the VGG-Sound dataset~\cite{chen2020vggsound} covering 309 categories. 
This ratio will undoubtedly increase when employing bigger batch sizes or fewer categories. 
% \jy{is there a more detailed number like 42.1 or something?} \jiayi{The probability is a statistical value, which fluctuates around 40\%.}
%
Secondly, we examine how the false negative issue might affect audio-visual localization performance.
%, which is the mainstream task for audio-visual representation learning.
%
Following prior works~\cite{chen2021localizing,mo2022localizing}, we adopt ResNet-18 as the backbone to encode audio and visual features and a standard NCE loss~\cite{oord2018representation} is used.
% to update model parameters. 
% Following the convention~\cite{senocak2018learning,chen2021localizing,mo2022localizing}, we adopt Consensus Intersection over Union (CIoU) as the evaluation metric (the higher the better).
% In particular, we manually control the data distribution during training to gradually increase the proportion of false negatives.
In particular, we manually substitute the true negatives with false negatives in the training samples. 
% \ZL{potential reviewer question: how do you control the distribution and how do you inject false negatives? Will this manual shift of distribution itself affect the performance?}
As shown in Fig~\ref{fig:pilot}, when the false negative rate is progressively increased, a significant performance decline is found, indicating that the model is perplexed by these similar samples. 
%\jy{better to state which task is it? Something like, self-supervised sound source localization, the mainstream task for audio-visual representation learning.}
%
% In addition, we manually eliminate {false negatives} during training to guarantee that all samples in the same batch come from distinct categories, resulting in improved performance.
% \wx{do we have this?} \jy{may need to provide quan or qual results.}
%
The experiments above demonstrate that false negatives substantially impact the model quality and cannot be disregarded.
% More experimental details are provided in supplementary materials.

\subsection{Mitigating False Negatives in Audio Visual Learning}\label{subsec:suppress}
% \jy{I changed the sec name from "False Negative Aware Audio-Visual Learning" to something like ""}
% As discussed,
% audio-visual contrastive learning  encodes two modalities in a unified latent space in the sense that all unpaired samples are negative to each other.
%
% However, 
% From the perspective above, 
% we validate that the unpaired positive samples, \textit{i.e.}, false negatives under this problem setting, are frequently encountered and may harm the representations. 

As discussed in~\cref{subsec:problem_def}, we prove that the unpaired positive samples (\textit{i.e.}, false negatives) existed in every mini-batch that may harm the representations. 
% To solve this problem, we hypothesize that potential false negatives can be mined in each modality without considering inter-modal correlations. 
Correspondingly, we propose to solve this problem by \name with two complementary methods, False Negatives Suppression (\methodone) and True Negatives Enhancement (\methodtwo). 
For the same goal of mitigating the false negative issue, \methodone identifies the false negative samples and regularizes to reduce their effects, while
\methodtwo enhances the contribution of true negatives by region-wise comparison, which also inadvertently suppresses false negatives.
Both methods can be seamlessly integrated into the audio-visual contrastive framework as regularization terms.

% We hypothesize that potential false negatives can be mined individually in each modality without considering inter-modal correlations. Based on this motivation, we propose FNS with two complementary methods. METHOD1 suppresses the misleading effects of false negatives. METHOD2 enhances the contribution of true negatives by region-wise learning. Both methods are seamlessly integrated into the audio-visual contrastive framework as regularization terms.

\begin{figure*}[!t]
   \begin{center}
   {\includegraphics[trim={0 3mm 0 0},clip,width=0.65\linewidth]{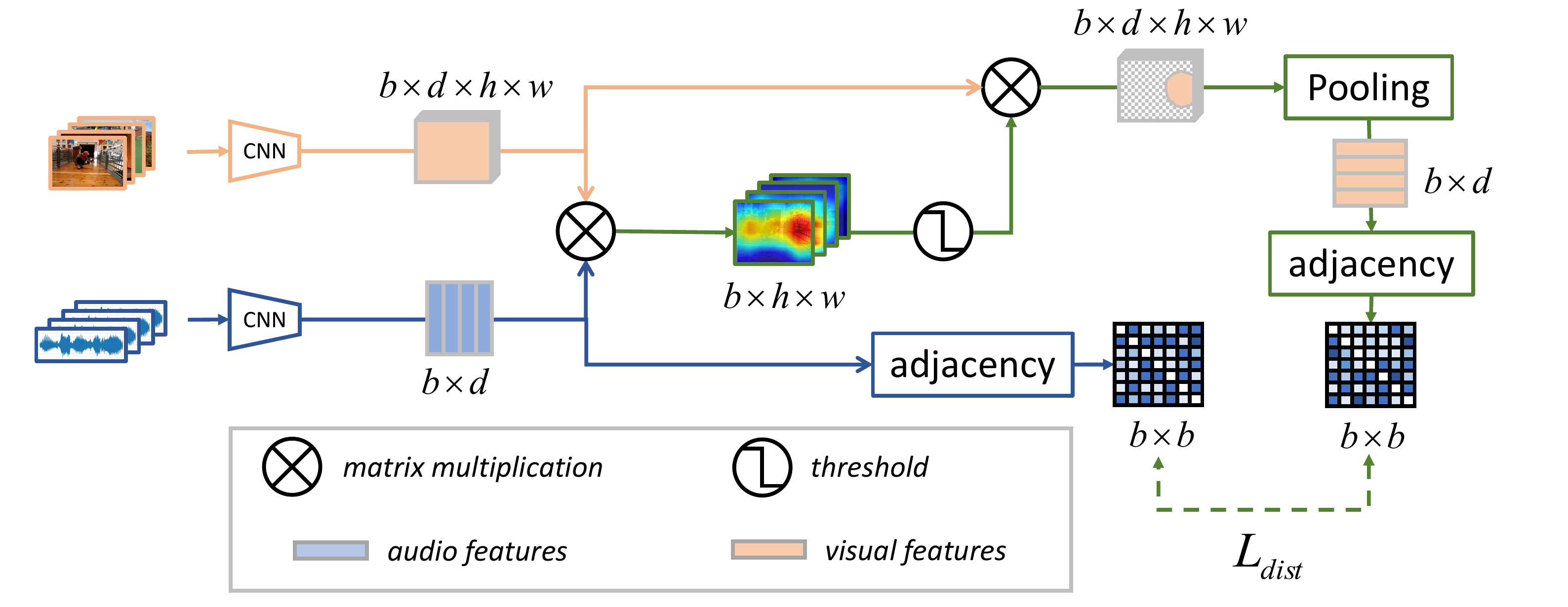}} 
   \end{center}
\caption{\textbf{An overview of True Negative Enhancement (\methodtwo). }The visual features from localized regions are extracted to construct a sound-source visual feature adjacency matrix. The audio adjacency matrix is used to regularize the sound-source visual feature adjacency matrix to enhance true negatives.}
\label{fig:overview region}
\vspace{-2pt}
\end{figure*}
\paragraph{\methodone: False Negatives Suppression.} To suppress the false negative effects, 
% the objective is to find out the false negative samples and eliminate their effects on the representation learning, which poses two challenges.
two challenges are posed.
First, distinguishing the potential false negative samples within the current mini-batch without extra supervision such as class labels. 
Second, eliminating the misleading effects of the identified false negatives.  

% Inspired by previous hard negative mining solutions~\cite{zheng2021weakly,chuang2020debiased} in the image self-supervision,
% Thus, during the audio-visual learning, we propose to directly calculate similarities within each modal without cross-modal interactions. 

For the first challenge, we propose to leverage the unimodal feature representations to calculate pair-wise sample similarities, \textit{i.e.}, adjacency matrix.
% Inspired by previous work~\cite{zheng2021weakly,chuang2020debiased,caron2018deep}, single-modal representations are widely used to measure similarities between two samples. 
% Inspired by the fact the feature similarity is more robust to obtain in single modality than cross modalities~\cite{zheng2021weakly,chuang2020debiased,caron2018deep}. 
% inspired by previous hard negative mining methods~\cite{zheng2021weakly,chuang2020debiased} in single modality, 
% Specifically, the single modality similarity  is more robust than 
As shown in Fig.~\ref{fig:overview}, 
the audio clips are fed into an audio encoder to obtain audio features  $Z^a \in \mathbb{R}^{b \times d}$, then we calculate a dot product with $(Z^a)^T$
followed by a row-wise softmax to obtain the pair-wise self-similarity matrix $S^a \in \mathbb{R}^{b \times b}$, \textit{i.e.}, the audio adjacency matrix.
Likewise, we average-pool the image features and obtain a visual adjacency matrix $S^v \in \mathbb{R}^{b \times b}$.
Such adjacency matrices encode pair-wise similarities across a batch without considering inter-modal connection.
% \jy{Is }
For each audio-visual sample, we show its nearest sample by querying the adjacency matrices, as shown in Fig~\ref{fig:similar samples}.
It indicates that intra-modal adjacency matrices can effectively identify potential class-matched samples. 
% The audio-visual pairs are input into a two-stream network and obtain audio features $Z^a \in \mathbb{R}^{b \times d}$ and visual features  $Z^v \in \mathbb{R}^{b \times d}$. 
% Then
% we directly calculate the pair-wise dot production followed by softmax to obtain the similarities between any two samples in audio and visual modalities respectively.
% we directly calculate two intra-modal adjacency matrices to measure the pair-wise similarities.
% For audio feature, we obtain its self-correlation matrix, i.e., adjacency matrix $M_a \in \mathbb{R}^{b \times b}$ by $Z^a \times (Z^a)^T$ followed by a row-wise softmax. 

% The adjacency matrix encodes pairwise similarities between any two audio clips. 
% Symmetrically, we obtain the adjacency matrix of images.
% The potential false negative samples are samples that have relatively higher other than the diagonal.
% Symmetrically, we obtain the visual adjacency matrix $M_v \in \mathbb{R}^{b \times b}$.  
% As shown in Fig..., we show the found similar samples which demonstrate the robustness of the intra-modal similarities.

% The second challenge is to eliminate the misleading effects of the identified false negatives. 
Regarding the second challenge, we propose to impose intra-modal adjacency as a soft supervision signal for inter-modal contrastive learning, \textit{i.e.}, audio-visual similarities should be statistically consistent with the intra-modal similarities. 
In other words, if two scenes are close, their audio features similarity value $\text{sim}(Z^a_i,Z^a_j)$ should be high, the visual similarity $\text{sim}(Z^v_i,Z^v_j)$ should match, and the cross-modal similarities $\text{sim}(Z^a_i,Z^v_j)$ or $\text{sim}(Z^v_i,Z^a_j)$ should also show consistency. %their audio feature similarity
% Given that learning the intra-modal similarity is much easier, 
Therefore, we leverage intra-modal similarity as a supervision signal for audio-visual contrastive learning.
%
% \jy{modify these to match your}
% In other words, consider two audio-visual pairs, if the two audio-clips are similar,  .......\wx{some intuitive thing here to help reader understand }
% In other words, if one audio clip is similar to another arbitrary audio clip in the batch, then this audio clip should also have similar feature with the 
% \NB{more intuitive also here - such as: i.e. if two images show the same object, then corresponding audio-visual pair also show the same object.}
Technically, we could utilize hard thresholds to achieve pseudo labels from intra-modal similarity like~\cite{senocak2022learning,chen2021localizing,zheng2021weakly}, which can serve to identify potential false negatives samples or areas.
However, such methods require parameter tuning, and misassigned labels can exacerbate the false-negative problem. 
% For instance, in the third column of Fig~\ref{fig:similar samples}. 
% Two pictures with man playing ukulele may generate different sounds
% although they are highly matched in the visual representation.
Differently, we propose
\methodone that directly regularizes the similarity scores. For a sample $i$, its optimization objective is formulated as:
%
%
% Previous work~\cite{zheng2021weakly} 
% Clustered pseudo labels~\cite{caron2018deep} or 
% Hard thresholds~\cite{chen2021localizing} or 
% Instead of involving hard thresholds to select potential hard negatives, we apply the adjacency matrices as a soft label to regularize the training.
% Specifically, we follow the convention to obtain the cross-modal contrastive matrix $M_c \in \mathbb{R}^{b \times b}$ by computing the matrix production of $Z^a$ and $Z^v$ and adopt the NCE loss~\cite{oord2018representation}. 
% Besides the NCE loss~\cite{oord2018representation} being unchanged, 
% Specifically, we still adopt the NCE loss and use the $M^a$ and $M^v$ as two extra regularization terms.
% The optimization objective is formulated as:
%
% \begin{eqnarray}
% \label{equation contra}
%     L_1 = \alpha\frac{1}{b} \sum_j^b||\frac{1}{\tau} {sim}(Z^a_i,Z^v_j) - \frac{1}{\tau} {sim}(Z^a_i,Z^a_j)||_1 \label{equation loss audio} \\
%     L_2 = \alpha\frac{1}{b} \sum_j^b||\frac{1}{\tau} {sim}(Z^a_i,Z^v_j) - \frac{1}{\tau} {sim}(Z^v_i,Z^v_j)||_1 \label{equation loss image}
% \end{eqnarray}
%
\begin{eqnarray}
% \label{equation contra}
    \mathcal{L}_{\methodone\_1} = \frac{1}{b}  \sum_j^b \mathcal{L}_{dist} (\text{sim}(Z^a_i,Z^v_j), \text{sim}(Z^a_i,Z^a_j)) \\
    \mathcal{L}_{\methodone\_2} = \frac{1}{b} \sum_j^b \mathcal{L}_{dist} (\text{sim}(Z^a_i,Z^v_j), \text{sim}(Z^v_i,Z^v_j))  \label{equation loss image}
\end{eqnarray}

% \jiayi{Maybe we could replace the concrete function with $L_{dist}$ and conduct a ablation study on different distance functions, \textit{e.g.}, L1, L2 and smooth L1, etc.}

% Intuitively, the diagonals of the adjacency matrices would dominant

% The $\text{sim}(Z^v_i,Z^v_j)$ and $\text{sim}(Z^a_i,Z^a_j)$ denote pair-wise similarities obtained by a row-wise $\text{softmax}$ function in each modality.
%  \NB{softmax applied row-wise to each modality? should you be more specific here or is it clear?}. 
 % $\alpha$ and $\beta$ are hyper parameters. 
%  \jy{$\alpha \frac{1}{b}$ looks like can be omitted. It can be adjusted by loss weights}
% As shown, we directly constrain the pair-wise similarities by calculating the l1 distances between the inter-modal adjacency matrix and intra-modal adjacency matrices.
For an audio sample $a_i$, the NCE loss $L_{\text{contrast\_i}}$ in Eq.~\ref{equation nce loss} pushes all negatives away by encouraging $\text{sim}(Z^a_i, Z^v_j)_{j \in[0,n), j \neq i}$ to be close to 0,
whereas \methodone pulls back false negatives to suppress their effects.
The suppression intensity corresponds to the values of $\text{sim}(Z^a_i, Z^a_j)$ and $\text{sim}(Z^v_i, Z^v_j)$. 
% \jy{need a name for sth like L1?}
In Eq.~\ref{equation loss image}, the visual contrastive loss is calculated symmetrically.
In practice, we calculate L1 distances between the inter-modal contrastive matrix and intra-modal adjacency matrices as shown in Fig.~\ref{fig:overview}, so all pair-wise similarities are considered.
Such a regularization term yields a parameter-free process so that we do not need to choose a hard threshold to determine the real false negatives, as the false negative effects can be adaptively regularized.

% 2). The original NCE loss~\cite{oord2018representation} is not modified, which ensures the lower bound to be still proportional to batch size. \jiayi{The Lower bound of NCE seems to have nothing to do with our contributions. Discuss later.}

% \subsection{\methodtwo: Adjacency as True Negatives Enhancer}
% \jiayi{The titles of section 3.2 and 3.3 are a little confusing.}

\paragraph{\methodtwo: True Negatives Enhancement.}\label{subsec:enhance}
% In this section, we further enhance the contribution of true negatives by region-wise comparison, to relatively reduce the misleading effect of false negatives. 
To further reduce the misleading effect of false negatives, we propose to enhance the contribution of true negatives by region-wise comparison.
% In most cases, a small region of the image contains the real sounding object, with the majority of the left areas being background or quiet objects~\cite{chen2021localizing}.
% Background can be well classified as it is far away from the objects in the latent space but the quiet objects are often falsely localized.
% Different sounds are intuitively produced by different regions/pixels.
% We are interested in region-wise comparison because.
As opponents, the impact of true negatives and that of false negatives are relative, so if the role of true negatives is raised, the role of false negatives would be suppressed.
%
% While, under the setting of audio-visual localization,
% Inspired by the fact that the in the setting of audio-visual contrastive learning, visual content 
% We notice that in audio-visual contrastive learning, 
To improve the effect of true negatives in audio-visual learning, we turn back to its core concept, \textit{i.e.}, that the \textit{sound source objects} are different both audibly and visually between the true negatives.
Therefore, it is straightforward to put a particular emphasis on the possible regions of genuine sound-emitting objects.
% instead of the whole image

% \wx{In previous methods, ~\cite{mo2022localizing} only considers the most similar regions between the audio and image and disregard rest areas. ~\cite{chen2021localizing} only consider the intra-image region comparing, no inter-image region comparing. shall we mention these differences
% }
%
Specifically, we localize the sound-source objects by paired audio-visual samples,
% aforementioned audio adjacency matrix
pop up their regional visual features, and encourage those of true negative samples to be pulled away.
The process is called true negative enhancement (TNE). 
It is worth noting that similar regularizations have been mentioned but disregarded by previous methods~\cite{mo2022localizing,chen2021localizing,senocak2022learning} because they did not distinguish the true negatives and false negatives samples.

\begin{table*}[ht!]
\centering
\caption{Quantitative results of the model trained with Flickr 10k and 144k. 
Note that 'EZ-VSL + OGL' corresponds to the main results reported in ~\cite{mo2022localizing}. 
'EZ-VSL’ 
indicates our reproduced results without OGL, which are not reported in ~\cite{mo2022localizing}. 
We reproduce the results with the trained weights and code provided by ~\cite{mo2022localizing}.
% EZ-VSL*: we reproduce the results with the trained weight and code provided by ~\cite{mo2022localizing}.
}
\centering\scalebox{.75}{
\setlength{\tabcolsep}{2.5mm}
\begin{tabular}{lccccc}
\toprule
Train set & Method & Flickr CIoU(\%) & Flickr AUC(\%) & VGG-SS CIoU(\%) & VGG-SS AUC(\%) \\
\hline
\multirow{8}{*}{Flickr 10k }
&Attention10k~\cite{senocak2018learning} & 43.60 & 44.90 & - & -  \\
&CoursetoFine~\cite{qian2020multiple} &52.20 & 49.60 & - & - \\
&AVObject~\cite{afouras2020self} &54.60 & 50.40 & - & - \\
&LVS~\cite{chen2021localizing} &58.20 & 52.50 & - & -  \\
&EZ-VSL*~\cite{mo2022localizing} &62.24 & 54.74 & 19.86 & 30.96 \\
&Ours & \textbf{84.33} & \textbf{63.26} & \textbf{35.27} & \textbf{38.00}  \\
\cline{2-6}
&EZ-VSL + OGL~\cite{mo2022localizing} &81.93 & 62.58 & 37.61 & 39.21 \\
&Ours + OGL & \textbf{84.73} & \textbf{64.34} & \textbf{40.97} & \textbf{40.38} \\
\hline
\multirow{8}{*}{Flickr 144k }
&Attention10k~\cite{senocak2018learning} &66.00 & 55.80 & - & -  \\
&DMC~\cite{hu2019deep} &67.10 &56.80 &- &- \\
&LVS~\cite{chen2021localizing} &69.90& 57.30 &-&-\\
&HardPos~\cite{senocak2022learning} & 75.20 & 59.70&-&- \\
&EZ-VSL*~\cite{mo2022localizing} & 72.69 & 58.70 & 30.27 & 35.92 \\
&Ours  & \textbf{78.71} & \textbf{59.33} & \textbf{33.93} & \textbf{37.29} \\
\cline{2-6}
&EZ-VSL + OGL~\cite{mo2022localizing} &83.13 & 63.06 & 41.01 & 40.23 \\
&Ours + OGL & \textbf{83.93} & \textbf{63.06} & \textbf{41.10} & \textbf{40.44} \\
\bottomrule
\end{tabular}}
\vspace{-5pt}
\label{table results trained on flickr}
\end{table*}

% Thus, we propose to use the audio-only similarities to guide the localization. 
We show the paradigm of \methodtwo in Fig.~\ref{fig:overview region}. 
Given an audio-visual pair $(a_i, v_i)$, 
% \NB{should there be brackets around the pair?}
we obtain its localization result and use the localization map as a mask to extract the localized visual representation $Z^s_i \in \mathbb{R}^{d \times h \times w}$, where \textit{s} indicates it is sounding region visual representation. 
%
% \jy{ "s" in $Z^s_i$ is the abbreviation of what} \wx{$Z^s_i$ is sounding region visual representation}
%
%
In other words, $Z^s_i$ denotes the visual features that are aligned with the paired audio features.
Then, consider another arbitrary audio-visual pair $(a_j, v_j)$ which localizes visual features $Z^s_j$. 
% If $a_i$ and $a_j$ are semantically matched, $Z^s_i$ should be similar to $Z^s_j$ which encourages the model
% to mine shared representation. 
If $a_i$ and $a_j$ are semantically different, i.e., $a_j$ is a true negative of $a_i$, then the $Z^s_i$ should be dissimilar to $Z^s_j$. 
It encourages the model to focus on different pixels so as to mine discriminative visual features according to the audio similarities.
To leverage such a constraint in practice, we regularize the audio adjacency matrix and the similarities between the sounding region visual features.
Formally, the \methodtwo regularization is:

% \begin{eqnarray}
% \label{equation contra}
%     L_3 = \frac{1}{b} \sum_j^b||\frac{1}{\tau} {sim}(Z^a_i,Z^a_j) - \frac{1}{\tau} {sim}(Z^s_i,Z^s_j)||_1
% \end{eqnarray}

\begin{eqnarray}
% \label{equation contra}
    \mathcal{L}_{\methodtwo} = \frac{1}{b} \sum_j^b \mathcal{L}_{dist} (\text{sim}(Z^a_i,Z^a_j), \text{sim}(Z^s_i,Z^s_j))
\end{eqnarray}

\methodone and \methodtwo are two general mechanisms for multi-modal contrastive learning.
\methodone adopts audio and visual adjacency matrices to explore the potential false negatives and suppress their misleading effects on the NCE loss.
\methodtwo uses audio adjacency to discriminate the sound-source localization so the model tends to discover the genuine sound sources.
Both methods can be seamlessly integrated with the existing contrastive learning framework as extra regularization terms.
Our final optimization objective for the $i$-th sample pair is:
\begin{eqnarray}
% \label{equation contra}
    \mathcal{L}_i = \mathcal{L}_{\text{contrast\_i}} + \alpha \mathcal{L}_{\methodone\_1} + \beta \mathcal{L}_{\methodone\_2} + \gamma \mathcal{L}_{\methodtwo}
\end{eqnarray}
where $\alpha,\beta,\gamma$ are hyperparameters.
% \jiayi{I'm a little confused about equation (5). L1, L2, L3 are not batch-wise loss? Maybe we could replace it with $L = L_{NCE} + L_1 + L_2 + L_3$ and correspondingly modify equation (1). }

% We first calculate the localized sounding regions of each audio-visual pair and adopt the localized maps to mask out background. 
% Then the sounding image representations $Z^v \in \mathbb{R}^{b \times d}$
% are extracted with average pooling.
% Following our intuition, the visual features are pushed apart guided by the potential true negatives provided by the audio adjacency.
% Formally, the \methodtwo regularization is denoted as:
% Given the audio adjacency matrix, the region-wise regularization can be achieved:

% where ...
% The NCE loss 

% As a result, the model is encouraged to exploit different pixels when different sounds are presented.
% Consequently, the quiet objects are ignored as they fail to provide discrimination between classes and genuine sounding objects are localized.
% Nevertheless, 
% such regularization implicitly clusters the samples into semantically matched groups in the latent space, which leads to high-quality audio-visual representations. 

\section{Experiments}
\label{section exp}
\subsection{Experimental Settings}
\paragraph{Datasets}
We train our audio-visual localization model on two datasets:
Flickr SoundNet~\cite{aytar2016soundnet} and VGG-Sound~\cite{chen2020vggsound}.
Flickr SoundNet contains 2 million unconstrained videos from
Flickr. For a fair comparison with the existing methods~\cite{mo2022localizing,chen2021localizing,senocak2022learning,liu2022exploiting}, we conduct the training on two subsets of 10k and 144k paired samples from Flicker SoundNet.
VGG-Sound includes 200k video clips from 309 classes. We also train on two subsets of 10k and 144k paired samples following the convention.

Localization performances are measured on four benchmarks, Flickr~\cite{aytar2016soundnet}, VGG-SS~\cite{chen2020vggsound}, Heard 110 and AVSBench~\cite{zhou2022avs}.
The Flickr test set has 250 audio-visual pairs with manually labeled bounding boxes.
VGG-SS is more challenging with 5,000 audio-visual pairs over 220 categories.
Heard 110 is another subset of VGG-Sound 
% which includes 70k training samples 
to test the open-set learning ability. 
Its train set has 110 classes and the \textit{val} set has another disjoint 110 unheard/unseen classes.
Finally, AVSBench~\cite{zhou2022avs} is a recently proposed audio-visual dataset with 5,356 videos over 23 classes. It provides pixel-wise labels for fine-grained localization evaluation.

% Heard 110~\cite{chen2021localizing} and AVSbench~\cite{zhou2022avs}.
% % \jiayi{I think we should introduce train sets and test sets separately in case of ambiguity. Moreover, the annotation in test sets shall be specified. }
% Flickr SoundNet contains 2 million unconstrained videos from
% Flickr. For a fair comparison with the existing methods~\cite{mo2022localizing,chen2021localizing,senocak2022learning,liu2022exploiting}, we conduct the training on two subsets of 10k and 144k paired samples from the Flicker SoundNet.
% VGG-Sound includes 200k video clips from 309 classes. We also train on two subsets of 10k and 144k paired samples following the convention.
% Heard 110 is another subset of VGG-Sound which includes 70k training samples to test the open-set learning ability. 
% Its train set has 110 classes and the validation set has another disjoint 110 unheard/unseen classes.
% Finally, AVSbench~\cite{zhou2022avs} is a recently proposed audio-visual dataset with xxx samples from xx classes. It provides pixel-wise labels for fine-grained localization evaluation.

\paragraph{Implementation Details}
We implement our method with PyTorch.
The images are resized and randomly cropped into $224 \times 224$ resolution, together with random horizontal flipping. 
The audio inputs are extracted from 3 seconds of audio clips and converted into log spectrogram maps. We also apply audio augmentation including Frequency mask and Time mask~\cite{park2019specaugment}.
For both visual and audio encoders, we adopt ResNet18~\cite{he2016deep} and the visual encoder is pre-trained on ImageNet-1k~\cite{deng2009imagenet}.
The model is optimized for 30 epochs with Adam using a learning rate of $10^{-4}$ and a weight decay of $10^{-4}$. To achieve a stable representation, we warm up the network with only NCE loss for 3 epochs, then integrate our regularization for the remaining epochs. 

% \paragraph{Post-processing: Object-guided Localization}

\begin{table*}[t]
\caption{
Quantitative results of models trained with VGG-SS 10k and 144k.
% EZ-VSL*: we reproduce the results with the trained weight and code provided by ~\cite{mo2022localizing}.
}
\centering
\centering\scalebox{.75}{
\setlength{\tabcolsep}{2.0mm}
\begin{tabular}{lccccc}
\toprule
Train set & Method & Flickr CIoU(\%) & Flickr AUC(\%) & VGG-SS CIoU(\%) & VGG-SS AUC(\%) \\
\hline
\multirow{5}{*}{VGGSound 10k}
&LVS~\cite{chen2021localizing} &61.80 & 53.60 & - & -   \\
&EZ-VSL*~\cite{mo2022localizing} &63.85 & 54.44 & 25.84 & 33.68 \\
&Ours &\textbf{85.74} & \textbf{63.66} & \textbf{37.29} & \textbf{38.99}  \\
\cline{2-6}
&EZ-VSL + OGL~\cite{mo2022localizing} &78.71 & 61.53 & 38.71 & 39.80 \\
&Ours + OGL & \textbf{82.13} & \textbf{63.64} & \textbf{40.69} & \textbf{40.42} \\

\hline
\multirow{9}{*}{VGGSound 144k}
&Attention10k~\cite{senocak2018learning} &- & - & 18.50 & 30.20  \\
&DMC~\cite{hu2019deep} & - & - & 29.10 & 34.80 \\
&AVObject~\cite{afouras2020self} & - & - & 29.70 & 35.70 \\
&LVS~\cite{chen2021localizing} & 73.50 & 59.00 & 34.40 & 38.20 \\
&HardPos~\cite{senocak2022learning} & 76.80 & 59.20 & 34.60 & 38.00 \\
&EZ-VSL~\cite{mo2022localizing} & 79.51 & 61.17 & 34.38 & 37.70 \\
&Ours  & \textbf{84.73} & \textbf{63.76} & \textbf{39.50} & \textbf{39.66} \\
\cline{2-6}
&EZ-VSL + OGL~\cite{mo2022localizing} &83.94 & 63.60 & 38.85 & 39.54 \\
&Ours + OGL & \textbf{85.14}  & \textbf{64.30} & \textbf{41.85} & \textbf{40.80} \\
\bottomrule
\end{tabular}}
\label{table results trained on vgg}
\end{table*}

% \subsection{Comparison with State-of-the-arts}
\subsection{Comparison on Flickr-SoundNet and VGG-SS}
\paragraph{Flickr-SoundNet}
We train our model on Flickr 10k and 144k and 
report performances in Table~\ref{table results trained on flickr}. 
\name achieves superior performances over previous methods on both Flickr and VGG-SS test sets.
% As shown, when tested on the Flickr test set, \name achieves significantly better performances than previous methods with both training sets (10k, 144k). 
Notably, 
compared to previous state-of-the-art EZ-VSL~\cite{mo2022localizing} on the Flickr test set, \name achieves a striking improvement of 22.09\% CIoU with 10k training samples and 6.02\% CIoU with 144k training samples.
When tested on the more challenging VGG-SS, \name also outperforms EZ-VSL by 15.41\% CIoU and 3.49\% CIoU respectively.
Finally, Object-Guided localization (OGL) is a post-processing strategy that adopts pure visual-based localization results to refine audio-visual localization.
For a fair comparison, we integrate \name with OGL to compare with EZ-VSL which also uses OGL. As shown, \name also outperforms EZ-VSL in most cases.
\paragraph{VGG-SS}
% \noindent\textbf{VGG-SS}
Performance of the model trained on VGG-Sound 10k and 144k is reported in Table~\ref{table results trained on vgg}.
\name also beats all previous methods 
% on both Flickr and VGG-SS test sets 
with a clear margin. For example, on the VGG-SS test set, we outperform EZ-VSL by 11.45\% CIoU with 10k and 5.31\% CIoU with 144k.

\begin{figure*}
\centering
\includegraphics[width=0.35\linewidth]{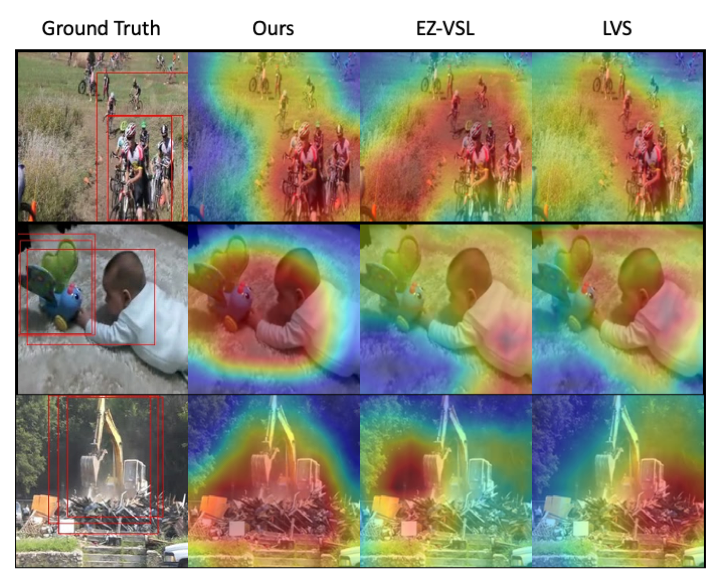}
\includegraphics[width=0.35\linewidth]{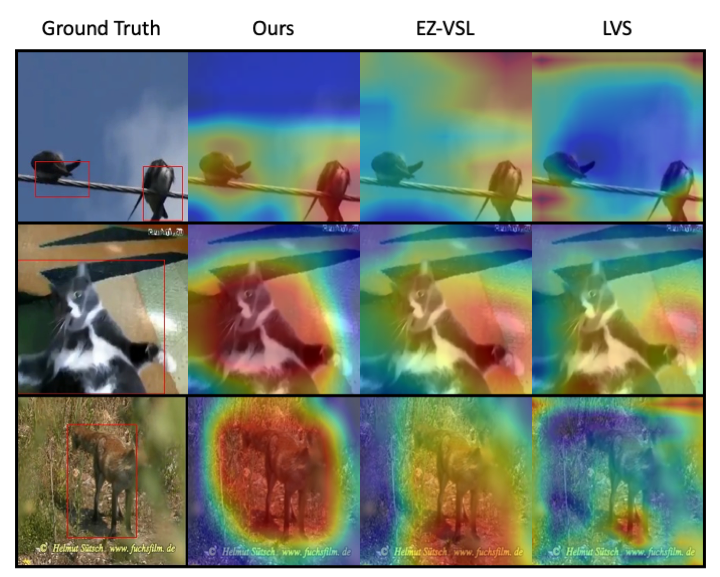}

\caption{Visualization comparison on Flickr-SoundNet (left) and VGG-SS (right) test sets.}
\label{fig: res}
\end{figure*}

We highlight two results as demonstrated in Table~\ref{table results trained on flickr} and Table~\ref{table results trained on vgg}.
First, \name achieves similar results on 10k and 144k training sets in both Flickr SoundNet and VGGSound, which is not observed in previous models.
We believe that small-scale datasets (Flickr 10k, VGG-Sound 10k) have fewer semantic classes and will encounter more false negatives during training, so previous methods have substantial performance gaps between the 10k and 144k training sets.
It indicates that \name can effectively address the false negative issue and has a strong ability for representation learning on small-scale datasets.
Second, we show cross-dataset evaluation results in both tables, i.e., train on Flickr and test on VGG-SS or vice versa. \name achieves strong results and outperforms existing methods, which  validates the cross-dataset generalization ability of \name. We show qualitative localization results in Fig.~\ref{fig: res}.

% \subsubsection{Comparison on Flickr-SoundNet and VGG-SS}
% In this section, we compare existing methods on two benchmarks: VGGSound-Source~\cite{chen2021localizing} and Flickr-SoundNet Test Set~\cite{senocak2018learning}. 
% Consistent with prior works, we report the results with different amounts of training samples (10k and 144k) on Flickr-SoundNet~\cite{aytar2016soundnet} and VGGsound~\cite{chen2020vggsound} in Tables~\ref{table results trained on flickr} and \ref{table results trained on vgg}, respectively.
% We observe substantial performance gains of CIoU and AUC compared with other methods on all train sets.
% It should be noted that our model significantly outperforms the current state-of-the-art methods by a large margin without the OGL module, \textit{e.g.}, CIoU improves 19.74\% and 7.22\% when trained on Flickr 10k and Flickr 144k, respectively. Such large performance gains indicate that our approach is able to better capture the audiovisual correspondence in videos, without introducing external visual priors. 

\subsection{Comparison on Heard 110 and AVSBench}
\paragraph{Heard 110}
To assess the generalization ability of \name in unseen/unheard audiovisual scenes, we conduct an open set experiment. 
% Following previous works \cite{chen2021localizing, mo2022localizing}, 
We
use the 70k samples covering 110 categories randomly sampled from VGGSound for training and then evaluate the model on the same 110 heard categories and another disjoint set with 110 unheard categories. 
As shown in Table~\ref{table openset}, \name considerably outperforms previous methods, especially on Unheard 110 (42.91\% \textit{vs.} 39.57\% of CIoU), which demonstrates the generalization ability of \name in unconstrained audio-visual data.

\begin{table}[t!]
\caption{Quantitative results on Heard 110 and Unheard 110. For a fair comparison, the results of EZ-VSL~\cite{mo2022localizing} and ours are integrated with the OGL module.}
\centering
\centering\scalebox{.8}{
\setlength{\tabcolsep}{2mm}
\begin{tabular}{lccc}
\toprule
Test Set & Method & CIoU(\%)  & AUC(\%) \\ \hline 
\multirow{3}{*}{Heard 110}
&LVS~\cite{chen2021localizing}&28.90&36.20\\
&EZ-VSL~\cite{mo2022localizing}&37.25&38.97\\
% &Ours &36.60&38.12\\
&Ours &\textbf{39.54}&\textbf{39.83} \\
\hline
\multirow{3}{*}{Unheard 110}
&LVS~\cite{chen2021localizing}&26.30&34.70\\
&EZ-VSL~\cite{mo2022localizing}&39.57&39.60\\
% &Ours & 39.29&39.45 \\
&Ours & \textbf{42.91} & \textbf{41.17} \\
\bottomrule
\end{tabular}}
\vspace{-2mm}
\label{table openset}
\end{table}

\paragraph{AVSBench}
AVSBench~\cite{zhou2022avs} is a newly proposed audio-visual segmentation benchmark with pixel-level annotations, which can be regarded as a fine-grained sound source localization task and used to accurately evaluate the localization ability of models. 
We directly perform a zero-shot evaluation on the AVSbench with metrics of mIoU and F-Score in two settings, Single Sound Source Segmentation (S4) and Multiple Sound Source Segmentation (MS3), without any fine-tuning. 
The results are reported in Table~\ref{table results on avsbench}. 
The proposed method achieves outstanding performance on both S4 and MS3 settings, \textit{e.g.}, 27.15\% and 21.98\% mIoU when trained on VGGSound 144k.
These results on the fine-grained localization benchmark validate the effectiveness of the \methodtwo, which enables the model to discriminate authentic sound-source regions.

% \begin{table*}[t]
% \caption{Quantitative results on AVSBench S4 and MS3~\cite{zhou2022avs}. 
% S4:Single Sound Source Segmentation. MS3:Multiple Sound Source Segmentation.}
% \centering
% \centering\scalebox{.85}{
% \setlength{\tabcolsep}{2.5mm}
% \begin{tabular}{cccccc}
% \hline
% Train set & Method & AVS\_S4 mIoU & AVS\_S4 FScore & AVS\_MS3 mIoU & AVS\_MS3 FScore \\
% \hline
% \multirow{3}{*}{Flickr 144k}
% &LVS &27.24 &.283 & 22.09 & .196   \\
% &EZ-VSL &21.19 & .249 & 14.74 & .158 \\
% &Ours &\textbf{25.13} & \textbf{28.19} & \textbf{18.74} & \textbf{19.00}  \\
% \hline
% \multirow{3}{*}{VGGSound 144k}
% &LVS &23.69 & .251 & 18.54 & .174 \\
% &EZ-VSL & 26.43 & .292 & 21.36 & .216 \\
% &Ours  &\textbf{27.15} &\textbf{.314} &\textbf{21.98} &\textbf{.225} \\
% \hline
% \end{tabular}}
% % \vspace{-5mm}
% \label{table results on avsbench}
% \end{table*}

\begin{table}[t!]
\centering
\caption{Zero-shot results on AVSBench S4 and MS3~\cite{zhou2022avs}. All models are pretrained on VGGSound-144k dataset. }
\scalebox{.8}{
\setlength{\tabcolsep}{5mm}
\begin{tabular}{lccc}
\toprule
Test set             & Method & mIoU  & FScore \\ \hline
\multirow{3}{*}{S4}  & LVS    & 23.69 & .251   \\
                     & EZ-VSL & 26.43 & .292   \\
                     & Ours   & \textbf{27.15} & \textbf{.314}   \\ \hline
\multirow{3}{*}{MS3} & LVS    & 18.54 & .174   \\
                     & EZ-VSL & 21.36 & .216   \\
                     & Ours   & \textbf{21.98} & \textbf{.225}   \\ \bottomrule
\end{tabular}
}
\vspace{-3mm}
\label{table results on avsbench}
\end{table}

\begin{table}[t!]
\caption{Analysis of each component of the proposed \name. aud adj: audio adjacency matrix. img adj: image adjacency matrix.}
\centering\scalebox{.75}{
\begin{tabular}{ccccc}
\toprule
\multicolumn{2}{c}{\methodone}        & \multirow{2}{*}{\methodtwo} & \multirow{2}{*}{Flickr CIoU(\%)} & \multirow{2}{*}{VGG-SS CIoU(\%)} \\ \cline{1-2}
\multicolumn{1}{c}{aud adj} & \multicolumn{1}{c}{img adj} &                          &                         &                      \\ \hline
              &                  &                      &      77.91                   &33.93      \\ %\hline
\checkmark  &&&79.91& 35.85 \\ %\hline
  &\checkmark&&81.92&36.58 \\ %\hline
\checkmark   &\checkmark&&84.33&36.92 \\ %\hline
 &&\checkmark& 81.12&36.04 \\ %\hline
\checkmark    &\checkmark & \checkmark&\textbf{85.74}&\textbf{37.29} \\
\bottomrule
\end{tabular}}
\vspace{-2mm}
\label{table ablation loss}
\end{table}

%%%%%%%%%%%%%%%%%%%%%%%%%%%%%%%%%%%%%%%%
% Move this table to supp. mat. (Table 7)
% \begin{table}[t!]
% \centering
% \caption{Ablation of different distance metrics for regularization loss. All models are trained on the VGG-Sound 10k training set.}
% \scalebox{.85}{
% \setlength{\tabcolsep}{4mm}
% \begin{tabular}{lcc}
% \toprule
% Loss & Flick CIoU(\%)  & VGGSS CIoU(\%) \\ \hline
% L2 & 80.72 & 34.02 \\
% Smooth L1 & 82.32 & 34.47 \\
% L1 &  85.74 & 37.29  \\
% \bottomrule
% \end{tabular}}
% \label{table ablate loss dist}
% \end{table}
%%%%%%%%%%%%%%%%%%%%%%%%%%%%%%%%%%%%%%%%

% \begin{table}[t!]
% \setlength{\tabcolsep}{2mm}
% \begin{tabular}{ccc|cc}
% \hline
% \multicolumn{2}{c}{\multirow{2}{*}{Multi-col-row}} & \multirow{2}{*}{Multi-col-row} & \multirow{2}{*}{Multi-col-row}  & \multirow{2}{*}{Multi-col-row} \\ 
% \hline
% & & 77.91 & 33.93   \\
% \checkmark &  & 84.33 & 36.92 \\
%  & \checkmark & 81.12  & 36.04 \\
% \checkmark&\checkmark & 85.74 & 37.29 \\ \hline
% \end{tabular}

% \caption{}

% \label{table ablate loss}
% \end{table}

% \begin{table}[t!]
% \setlength{\tabcolsep}{2mm}
% \begin{tabular}{cc|cc}
% \hline
% image adjacency & audio adjacency & Flickr CIoU  & VGG CIoU \\ \hline
% & & 77.91 & 33.93  \\
% \checkmark & & 81.92 &  36.58 \\
%  & \checkmark & 79.91 & 35.85  \\
% \checkmark&\checkmark &  84.33 & 36.92 \\ \hline
% \end{tabular}

% \caption{}

% \label{table ablate image audio}
% \end{table}

% \subsection{Ablation Analysis}
\subsection{Ablation Analysis  of \name}
We propose different regularization terms to guide audio-visual contrastive learning. 
In Table~\ref{table ablation loss}, we ablate each component individually. 
All results are obtained following the same hyperparameter setting on VGG-SS 10k and the baseline is trained with only NCE loss.
First, the audio adjacency matrix and visual adjacency matrix are obtained in \methodone to suppress the false negatives. 
We show that each adjacency matrix can improve the performances over the baseline, which indicates that potential false negatives can be effectively suppressed in both audio and visual modalities. By combining the two adjacency matrices together, \methodone achieves 84.33\% CIoU on Flickr and 36.92\% CIoU on VGG-SS.
Second, by only deploying \methodtwo, we achieve improved results over baseline (81.12\% CIoU on Flickr and 36.04\% CIoU on VGG-SS), indicating the effectiveness of \methodtwo in enhancing true negatives.
Finally, the combination of \methodone and \methodtwo achieves significant improvement over the baseline, showing the effectiveness of \name. 
We refer readers to supp. mat. for further ablation studies.

\subsection{Mining the Potential False Negatives}
\begin{figure}[t]
  \centering
  %\fbox{\rule{0pt}{2in} \rule{0.9\linewidth}{0pt}}
   \includegraphics[width=.8\linewidth]{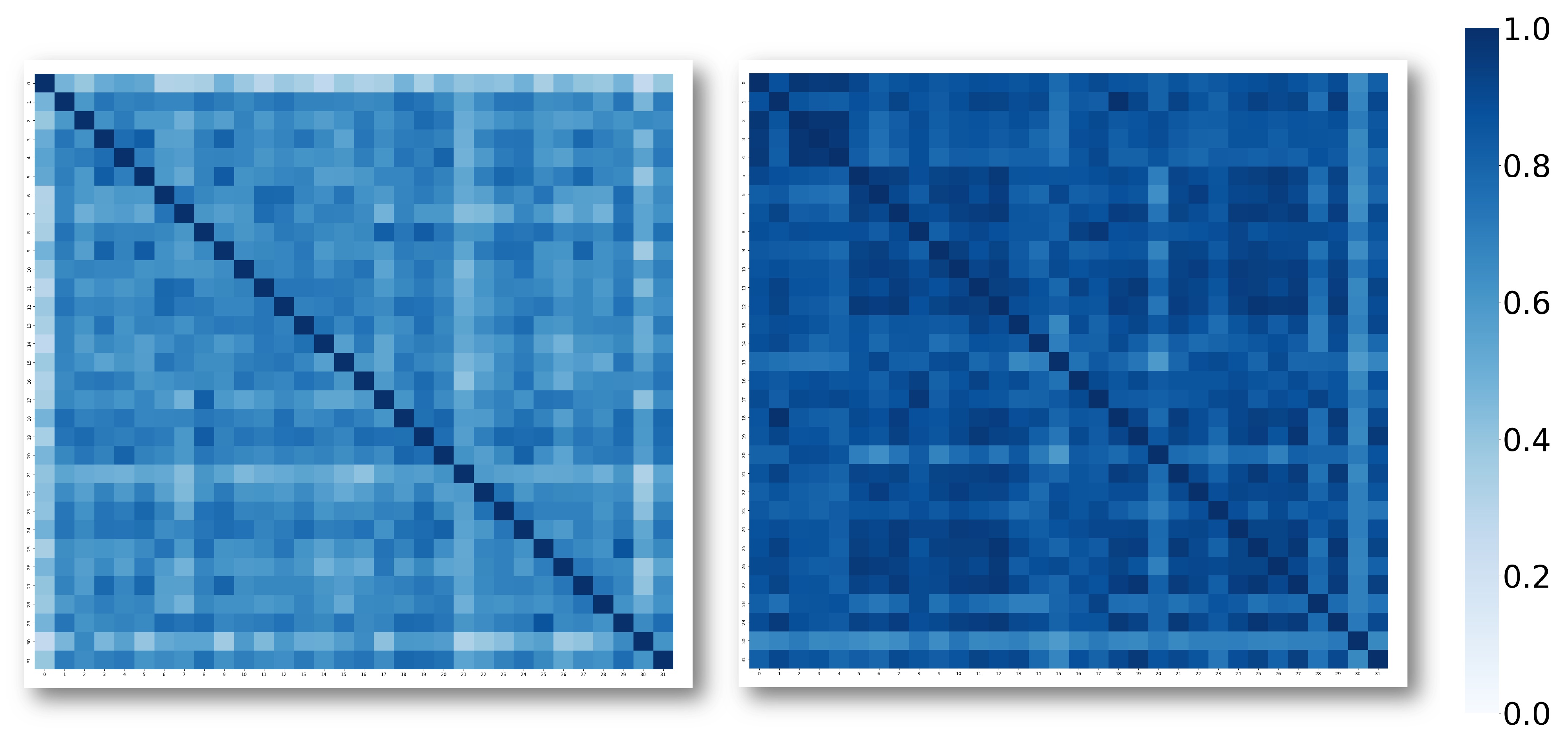}
   \caption{Cross-modal similarity matrix predicted by EZ-VSL (left) and ours (right) when all samples in the batch belong to the same category, \textit{namely}, they are false negatives of each other. All values are normalized between 0 to 1.}
   \label{fig:corr}
\end{figure}

In this section, we show quantitative and schematic analysis of the false negative mining ability of our method.
Intuitively, a good audio-visual model should be able to generate similar feature representations for class matched samples.
To validate this intuition, we construct a batch where all samples belong to the same category, i.e., all false negatives, and show the audio-visual similarity matrices of EZ-VSL~\cite{mo2022localizing} and ours in Fig.~\ref{fig:corr}.
As shown in the left, EZ-VSL only highlights the diagonal, 
since each audio feature is only similar to the paired image feature while all others are dissimilar despite all samples being from the same class.
Differently, the similarity distribution of our matrix is more evenly spread as the majority of the samples are regarded as similar. 
It indicates that \name has implicitly learned semantically-aware features and clustered them in the latent space, so false negatives will be effectively identified as they are closer in terms of feature distance.

Further, we show quantitative results of audio-visual similarities
in Table~\ref{table identify fn}
% which align with our observation
. When the batch contains all true negatives, our average similarity score is lower than previous methods~\cite{mo2022localizing,chen2021localizing}.
When the batch contains all false negatives, our average similarity score is higher.
The margin between the two similarities demonstrates our ability to distinguish false negatives and true negatives.

\begin{table}[t!]
\centering
\caption{Audio-visual similarities with different data.
TN: all samples in the batch belong to different categories.
FN: all samples in the batch belong to the same category.
% True Negative represents that all samples in the batch belong to different categories, while False Negative indicates the opposite, \textit{i.e.}, all belong to the same category.
}
\scalebox{.85}{
\setlength{\tabcolsep}{9mm}
\begin{tabular}{lcc}
\toprule
Method & TN $\downarrow$  & FN $\uparrow$ \\ \hline
LVS & 0.4484 & 0.5102 \\
EZ-VSl & 0.5858 &  0.5938 \\
Ours &  \textbf{0.3812}  &  \textbf{0.6554} \\
\bottomrule
\end{tabular}}
\vspace{-3mm}
\label{table identify fn}
\end{table}
%%%%%%%%%%%%%%%%%%%%%%%%%%%%%%%%%%%%%%%%
% Move to supp. mat.
% \subsection{Effect of Regularization Loss}
% In our implementation, we measure the distance between the adjacency matrices to enforce regularization.
% In Table~\ref{table ablate loss dist}, we ablate different distance metrics.
% As shown, using L1 distance achieves the best performance. 

% \subsubsection{Effect of Batch Sizes}
% In contrastive learning, larger 
%%%%%%%%%%%%%%%%%%%%%%%%%%%%%%%%%%%%%%%%

% \subsubsection{Effect of foreground threshold}

% \subsubsection{Effect of $\alpha$}

% \subsubsection{Two projectors}

% \subsection{Qualitative Results}

\begin{figure}
\centering
\includegraphics[width=0.8\linewidth]{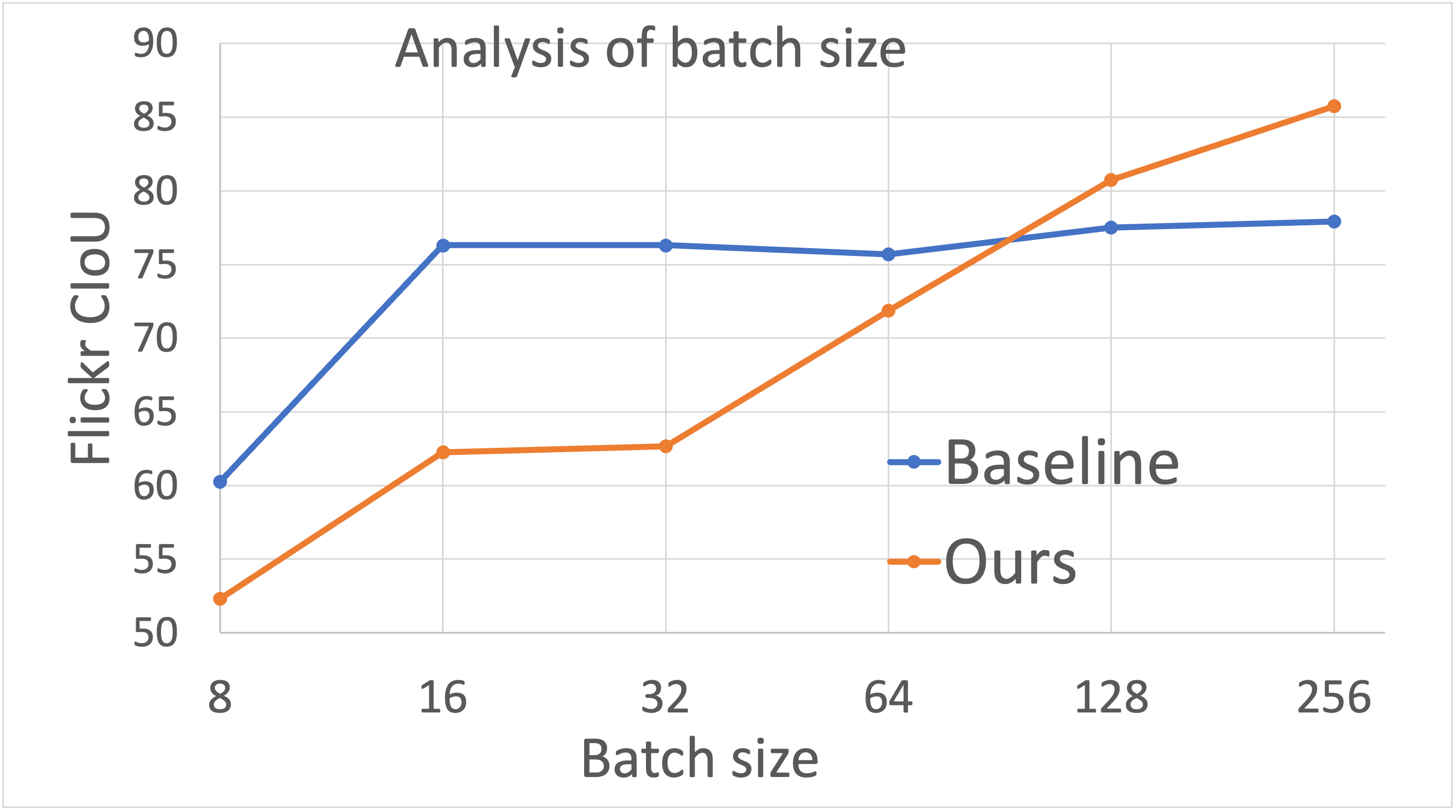}
% \vspace{-2mm}
\caption{Analysis of batch size on Flick. Our method boosts performances when batch size increases.}
% \vspace{-6mm}
\label{fig: batch size 1}
\end{figure}

\begin{figure}
\centering
\includegraphics[width=0.8\linewidth]{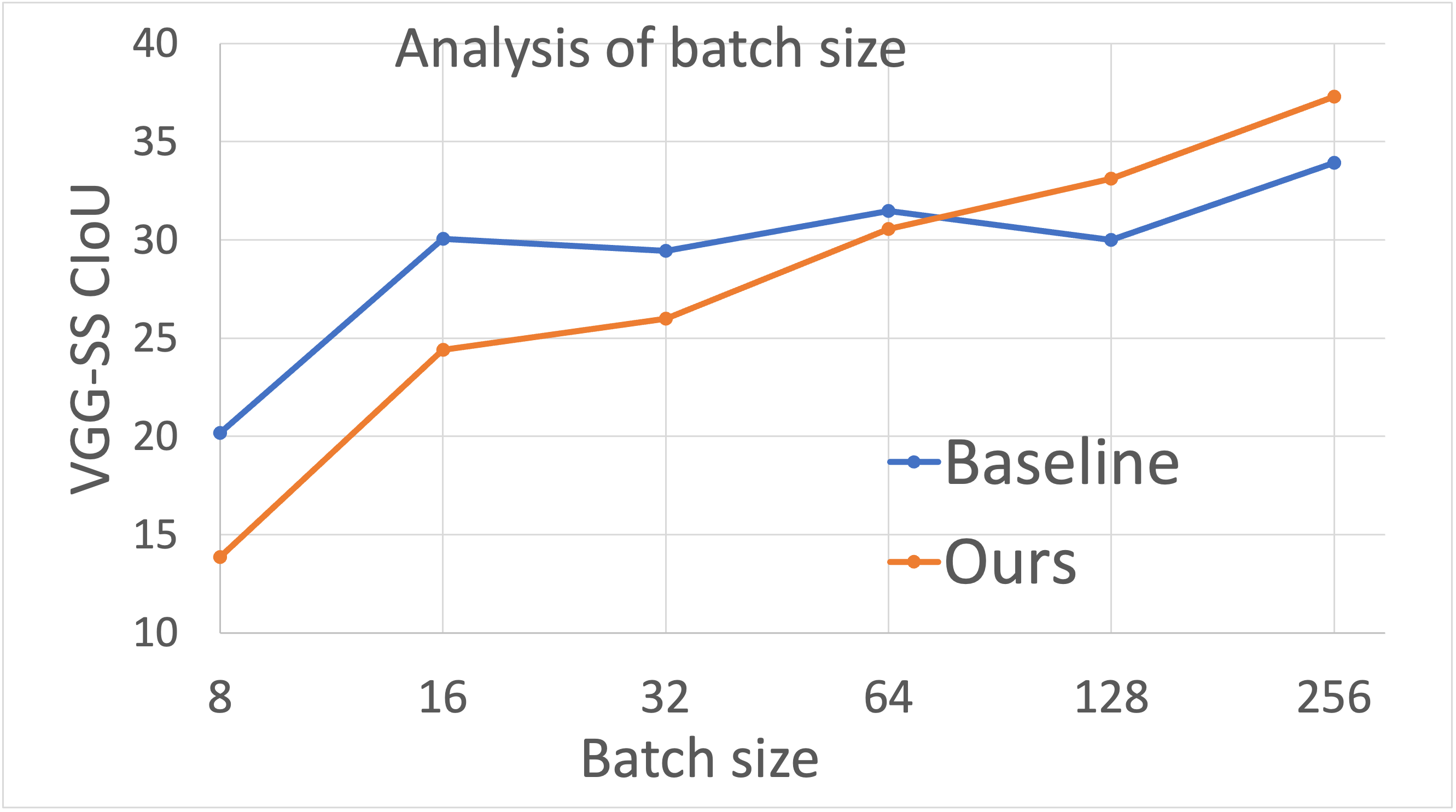}
% \vspace{-2mm}
\caption{Analysis of batch size on VGG-SS. Our method boosts performances when batch size increases.}
% \vspace{-6mm}
\label{fig: batch size 2}
\end{figure}

\subsection{Analysis of Batch Size}
In this section, we investigate the effects of batch sizes.
We show results with different batch size in Fig.~\ref{fig: batch size 1} and ~\ref{fig: batch size 2}. 
% \jiayi{Hi Weixuan, Fig.~\ref{fig: batch size 1} and \ref{fig: batch size 2} are somewhat redundant. Could we merge them into one figure?}
As shown, when the batch size is 64 or less, the false negative ratio is low and nearly insignificant, so our method provides no benefit or even hinders performance.
On the other hand, 
we observe a slight improvement in the baseline as batch size increases.
Nevertheless, false negatives become prominent in large batch sizes, which is the focus of our paper. 
Our proposed method improves performance by effectively addressing false negatives, outperforming the baseline by a clear margin.

\section{Conclusion}
In this paper, we propose a simple yet effective strategy named \name
to deal with the false negative issue in audio-visual sound source localization. 
We propose two complementary strategies to suppress false negatives (\methodone) and enhance true negatives (\methodtwo).
The intra-modal adjacency matrices of audio and visual are generated to identify false negatives.
Then two regularization terms are seamlessly incorporated with contrastive learning, which enable the model to learn semantic-aware features from audio-visual pairs.
We show that \name can effectively mitigate the false negative issues and achieve state-of-the-art performance on several audio-visual localization benchmarks.
We hope that our method can facilitate future research in audio-visual learning and other multi-modal tasks.

% \begin{figure*}
% \centering
% \subfigure{
% \begin{minipage}[t]{1\linewidth}
%     \includegraphics[width=0.48\linewidth]{cvpr2023-author_kit-v1_1-1/latex/figure/viz_flickr.pdf}
% \caption{图片1名称}
% \label{0}
% \end{minipage}
% }
% \subfigure{
% \begin{minipage}[t]{1\linewidth}
%     \includegraphics[width=0.48\linewidth]{cvpr2023-author_kit-v1_1-1/latex/figure/viz_vggss.pdf}
% \caption{图片2名称}
% \label{2}
% \end{minipage}
% }
% \caption{Qualitative results.}
% \end{figure*}

%%%%%%%%% REFERENCES
{\small
\bibliographystyle{ieee_fullname}
\bibliography{main}
}

\end{document}